\documentclass{article}

\usepackage{microtype}
\usepackage{graphicx}
\usepackage{subfigure}
\usepackage{booktabs} 
\usepackage{caption}
\usepackage{subfiles}

\usepackage{hyperref}

\usepackage{amsmath,amsfonts,bm}

\def\eqref#1{equation~\ref{#1}}

\def\floor#1{\lfloor #1 \rfloor}
\def\1{\bm{1}}

\def\va{{\bm{a}}}

\def\vw{{\bm{w}}}
\def\vx{{\bm{x}}}

\DeclareMathAlphabet{\mathsfit}{\encodingdefault}{\sfdefault}{m}{sl}
\SetMathAlphabet{\mathsfit}{bold}{\encodingdefault}{\sfdefault}{bx}{n}

\newcommand{\R}{\mathbb{R}}

\usepackage[accepted]{icml2021}

\icmltitlerunning{Statistical Measures For Defining Curriculum Scoring Function}

\begin{document}

\twocolumn[
\icmltitle{Statistical Measures For Defining Curriculum Scoring Function}



\icmlsetsymbol{equal}{*}

\begin{icmlauthorlist}
\icmlauthor{Vinu~Sankar Sadasivan}{to}
\icmlauthor{Anirban Dasgupta}{to}
\end{icmlauthorlist}

\icmlaffiliation{to}{Department of Computer Science \& Engineering, Indian Institute of Technology Gandhinagar, Gujarat, India}

\icmlcorrespondingauthor{Vinu~Sankar Sadasivan}{vinu.sankar@alumni.iitgn.ac.in}

\icmlkeywords{Machine Learning, ICML}

\vskip 0.3in
]



\printAffiliationsAndNotice

\begin{abstract}
Curriculum learning is a training strategy that sorts the training examples by some measure of their difficulty and gradually exposes them to the learner to improve the network performance. 
Motivated by our insights from implicit curriculum ordering, we first introduce a simple curriculum learning strategy that uses statistical measures such as standard deviation and entropy values to score the difficulty of data points for real image classification tasks. We empirically show its improvements in performance with convolutional and fully-connected neural networks on multiple real image datasets. We also propose and study the performance of a dynamic curriculum learning algorithm. Our dynamic curriculum algorithm tries to reduce the distance between the network weight and an optimal weight at any training step by greedily sampling examples with gradients that are directed towards the optimal weight. Further, we use our algorithms to discuss why curriculum learning is helpful. 
\end{abstract}

\section{Introduction}
\label{introduction}

Stochastic Gradient Descent (SGD) \citep{sgd} is a simple yet widely used algorithm for machine learning optimization. There have been many efforts to improve its performance. A number of such directions, such as AdaGrad \citep{adagrad}, RMSProp \citep{rmsprop}, and Adam \citep{adam}, improve upon SGD by fine-tuning its learning rate, often adaptively. However,~\citet{sgdbetter} has shown that the solutions found by adaptive methods generalize worse even for simple overparameterized problems. \citet{amsgrad} introduced AMSGrad hoping to solve this issue. Yet there is performance gap between AMSGrad and SGD in terms of the ability to generalize~\citep{switchsgd}. 
Hence, SGD still remains one of the main workhorses of the machine learning optimization toolkit.

SGD proceeds by stochastically making unbiased estimates of the gradient on the full data \citep{stochasticsampling}. However, this approach does not match the way humans typically learn various tasks. We learn a concept faster if we are presented  easy examples first and then gradually exposed to examples with more complexity, based on a curriculum. An orthogonal extension to SGD \citep{theoryofcl}, that has some promise in improving its performance is to choose examples according to a specific strategy, driven by cognitive science -- this is curriculum learning (CL) \citep{cl}, wherein the examples are shown to the learner based on a curriculum.

In this work, we propose two novel approaches for CL -- a practical algorithm that uses statistical measures for scoring examples and setting up curricula for image datasets, and a toy dynamic curriculum learning (DCL) framework. We do a thorough empirical study of our practical algorithm and provide some more insights into why CL works. Our contributions are as follows:

 \begin{itemize}
 
  \item We introduce a \textbf{simple, novel, and practical CL approach for image classification tasks that does the ordering of examples in an unsupervised manner using statistical measures}. Our insight is that statistical measures could have an association with implicit curricula ordering. We empirically analyze our argument of using statistical scoring measures (especially standard deviation) over combinations of multiple datasets and networks. 
    
    \item We propose a novel dynamic curriculum learning (DCL) algorithm to study the behaviour of CL. DCL is not a practical CL algorithm since it requires the knowledge of a reasonable local optima to compute the gradients of the full data after ever training epoch. DCL uses the \textbf{gradient information to define a curriculum that minimizes the distance between the current weight and a desired local minima after every epoch}. However, this simplicity in the definition of DCL makes it easier to analyze its performance formally.
    
     \item Our DCL algorithm generates a natural ordering for training the examples. Previous CL works have demonstrated that exposing a part of the data initially and then gradually exposing the rest is a standard way to setup a curriculum. We use two variants of our DCL framework to show that \textbf{it is not just the subset of data which is exposed to the model that matters, but also the ordering within the data partition that is exposed}. 
     
     \item We analyze how \textbf{DCL is able to serve as a regularizer and improve the generalization of networks}. Additionally, we study why CL based on standard deviation scoring works using our DCL framework. 

 \end{itemize}

\begin{figure}[t]
  \includegraphics[width=1\linewidth]{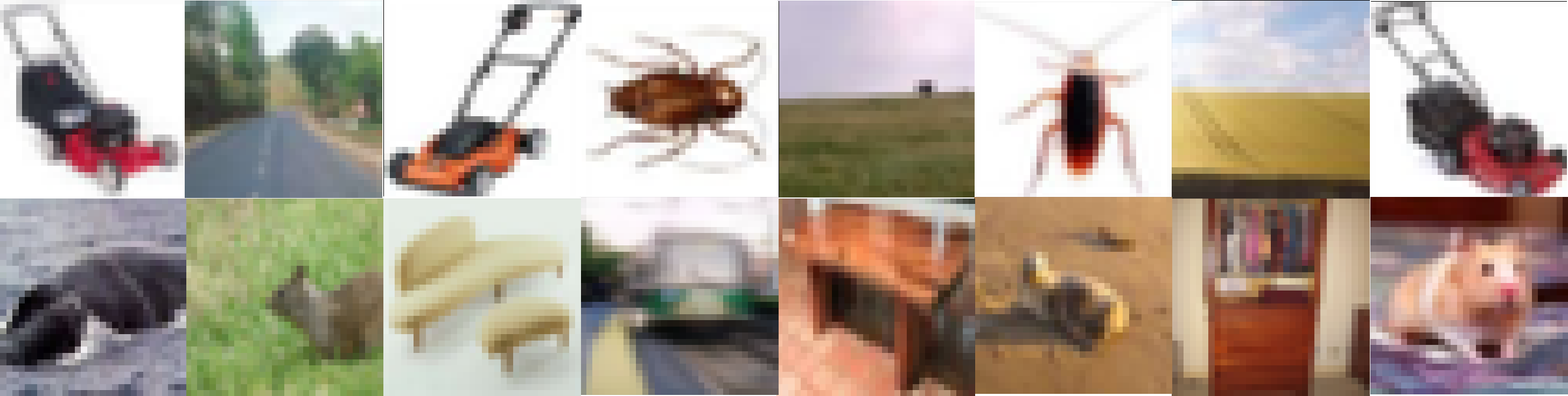}
  \includegraphics[width=1\linewidth]{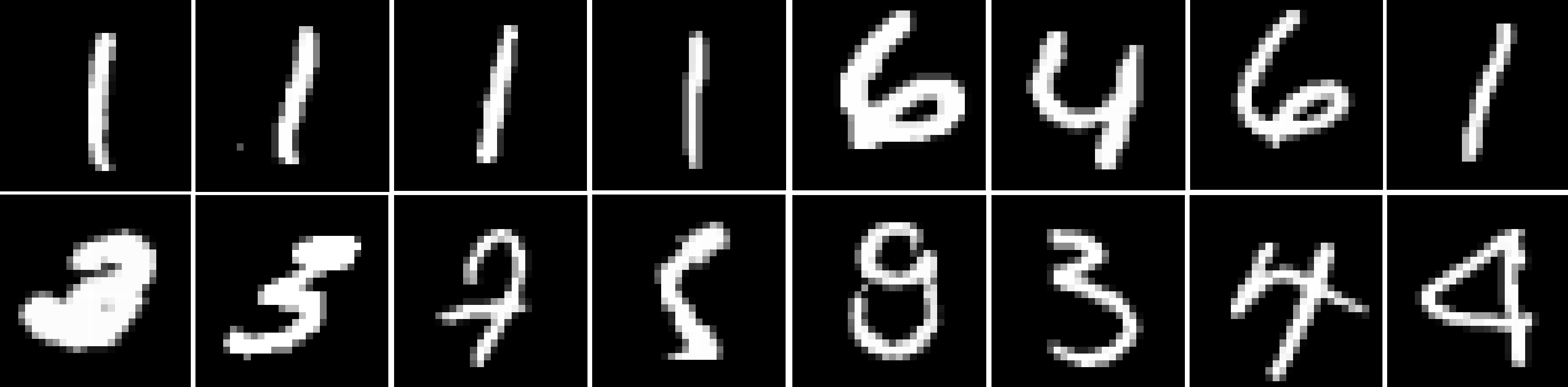}
  \vspace{-.1in}
  \caption{Implicit curricula: Top and bottom rows contain images that are learned at the beginning and end of the training, respectively. Top rows: CIFAR-100, bottom rows: MNIST.}
  \label{fig:implicitcurricula}
  \vspace{-.1in}
\end{figure}

\subsection{Related Works}
\label{relatedworks}

\citet{cl} formalizes the idea of CL in machine learning framework where the examples are fed to the learner in an order based on its \textit{difficulty}. The notation of difficulty scoring of examples has not really been formalized and various heuristics have been tried out: \citet{cl} uses manually crafted  scores, self-paced learning (SPL) \citep{spl} uses the loss values with respect to the learner's current parameters, and CL by transfer learning \citep{powerofcl} uses the loss values with respect to a pre-trained model to rate the difficulty of examples in a dataset. Among these works, what makes SPL particular is that they use a dynamic CL strategy, i.e., the preferred ordering is determined dynamically over the course of the optimization. However, SPL does not really improve the performance of deep learning models, as noted in \citep{learntoteach}. Similarly, \citet{ranking} uses a function of rank based on latest loss values for online batch selection for faster training of neural networks. \citet{impsample1} and \citet{impsample2} perform importance sampling to reduce the variance of stochastic gradients during training.  \citet{autocl} and \citet{tscl} propose teacher-guided automatic CL algorithms that employ various supervised measures to define dynamic curricula. The most recent works in CL show its advantages in reinforcement learning \citep{rl1, rl2}.

The recent work by \citet{theoryofcl} introduces the notion of \textit{ideal difficulty score} to rate the difficulty of examples based on their loss values with respect to a set of optimal hypotheses. They theoretically show that for linear regression, the expected rate of convergence at a training step for an example monotonically decreases with its ideal difficulty score. This is practically validated by \citet{powerofcl} by sorting the training examples based on the performance of a network trained through transfer learning. They also show that anti-curriculum learning, exposing the most difficult examples first, leads to a degrade in the network performance. However, there is a lack of theory to show that CL improves the performance of a completely trained network. Thus, while CL indicates that it is possible to improve the performance of SGD by a judicious ordering, both theoretical insights as well as concrete empirical guidelines to create this ordering remain unclear. In contrast to CL \citep{powerofcl}, anti-curriculum learning \citep{acl1, acl2, acl3} can be better than CL in certain settings.

\citet{letsagree} and \citet{whencl} investigate \textit{implicit curricula} and observe that networks learn examples in a dataset in a highly consistent order. Figure \ref{fig:implicitcurricula} shows the implicit order in which a convolutional neural network (CNN) learns data points from MNIST and CIFAR-100 datasets. \citet{whencl} also shows that CL (\textit{explicit curriculum}) can be useful in scenarios with limited training budget or noisy data. \citet{coreset} uses a coreset construction method to dynamically expose a subset of the dataset to robustly train neural networks against noisy labels. 

While the previous CL works employ tedious methods to score the difficulty level of the examples, \citet{audiovisualcl} uses the number of audio sources to determine the difficulty for audiovisual learning. \citet{normnmtcl} uses the norm of word embeddings as a difficulty measure for CL for neural machine translation. In light of these recent works, we discuss the idea of using statistical measures to score examples making it easy to perform CL on real image datasets without the aid of any pre-trained network.

\section{Preliminaries}
\label{preliminaries}

At any training step $\displaystyle t$, SGD updates the current weight $\displaystyle \vw_t$ using $\displaystyle \nabla f_i(\vw_t)$ which is the gradient of loss of example $\displaystyle \vx_i$ with respect to the current weight. The learning rate and the data are denoted by $\displaystyle \eta$ and $\displaystyle \mathcal{X} = \{(\vx_i,y_i)\}_{i=0}^{N-1}$, respectively, where $\displaystyle \vx_i \in [-1,1]^d$ denotes an example and $\displaystyle y_i \in [K]$ its corresponding label for a dataset with $\displaystyle K$ classes. Without loss of generality, we assume that the dataset is normalized such that $\sum_{i=0}^{N-1} \vx_i = \mathbf{0}$. We denote the learner as $\displaystyle h_\vartheta : [-1,1]^d \to [K]$. Generally, SGD is used to train $\displaystyle h_\vartheta$ by giving the model a sequence of mini-batches $\displaystyle \{B_0, B_1, ..., B_{T-1}\}$, where $\displaystyle B_i \subseteq \mathcal{X} ~\forall i \in [T]$. Traditionally, each $\displaystyle B_i$ is generated by uniformly sampling examples from the data. We denote this approach as \textit{vanilla}.

In CL, the curriculum is defined by two functions, namely the scoring function and the pacing function. The scoring function, $\displaystyle score_\vartheta(\vx_i, y_i) : [-1,1]^d \times [K] \to \R$, scores each example in the dataset. Scoring function is used to sort $\displaystyle \mathcal{X}$ in an ascending order of difficulty. A data point $\displaystyle (\vx_i,y_i)$ is said to be easier than $\displaystyle (\vx_j,y_j)$ if $\displaystyle score_\vartheta(\vx_i, y_i) < score_\vartheta(\vx_j, y_j)$, where both the examples belong to $\displaystyle \mathcal{X}$. Unsupervised scoring measures do not use the data labels to determine the difficulty of data points. The pacing function, $\displaystyle pace_\vartheta(t): [T] \to [N]$, determines how much of the data is to be exposed at a training step $\displaystyle t \in [T]$.

\begin{figure}[t]
  \includegraphics[width=1\linewidth]{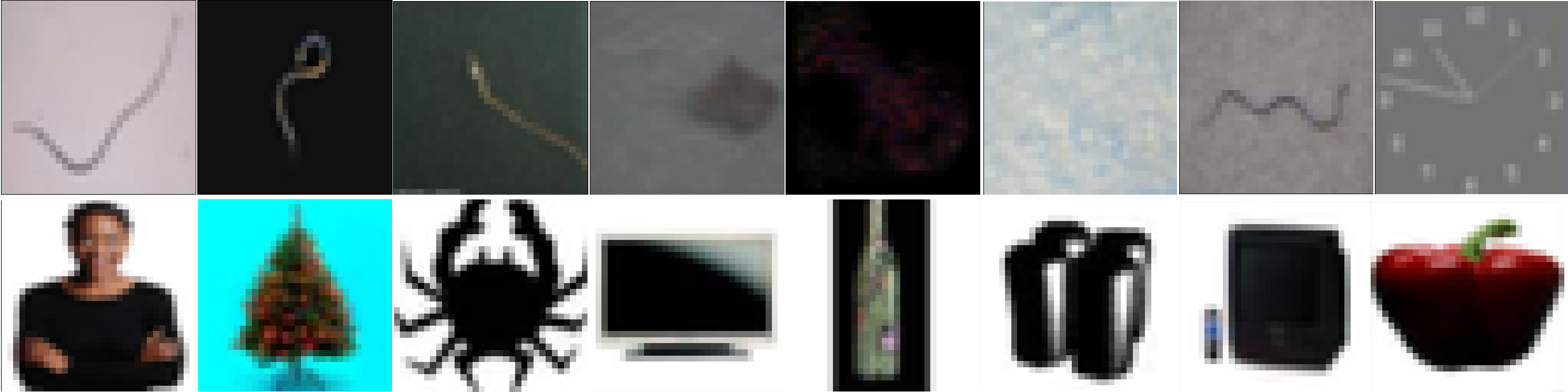}
  \vspace{-.1in}
  \caption{Top $\displaystyle 8$ images with the lowest standard deviation values (top row) and top $\displaystyle 8$ images with the highest standard deviation values (bottom row) in CIFAR-100 dataset.}
  \label{fig:stddevimages}
\end{figure}

\begin{algorithm}[tb]
   \caption{Curriculum learning method.}
   \label{algo:std}
\begin{algorithmic}
   \STATE {\bfseries Input:} Data $\displaystyle \mathcal{X}$, batch size $\displaystyle b$, number of mini-batches $T$, scoring function $\displaystyle score$, and pacing function $\displaystyle pace$.
   \STATE {\bfseries Output:} Sequence of mini-batches $\displaystyle [B_0, B_1, ..., B_{T-1}]$.
   \STATE sort $\displaystyle \mathcal{X}$ according to $\displaystyle score$, in ascending order.
   \STATE $\displaystyle B \gets [~]$
   \FOR{$i=0$ {\bfseries to} $T-1$}
   \STATE $\displaystyle size \gets pace(i)$
   \STATE $\displaystyle \tilde{\mathcal{X}}_i \gets \mathcal{X}[0,1,...,size-1]$
   \STATE uniformly sample $\displaystyle B_i$ of size $\displaystyle b$ from $\displaystyle \tilde{\mathcal{X}}_i$
   \ENDFOR
   \STATE {\bfseries return} $B$
\end{algorithmic}
\end{algorithm}

\begin{figure*}[t]
\centering
  \includegraphics[width=0.8\linewidth]{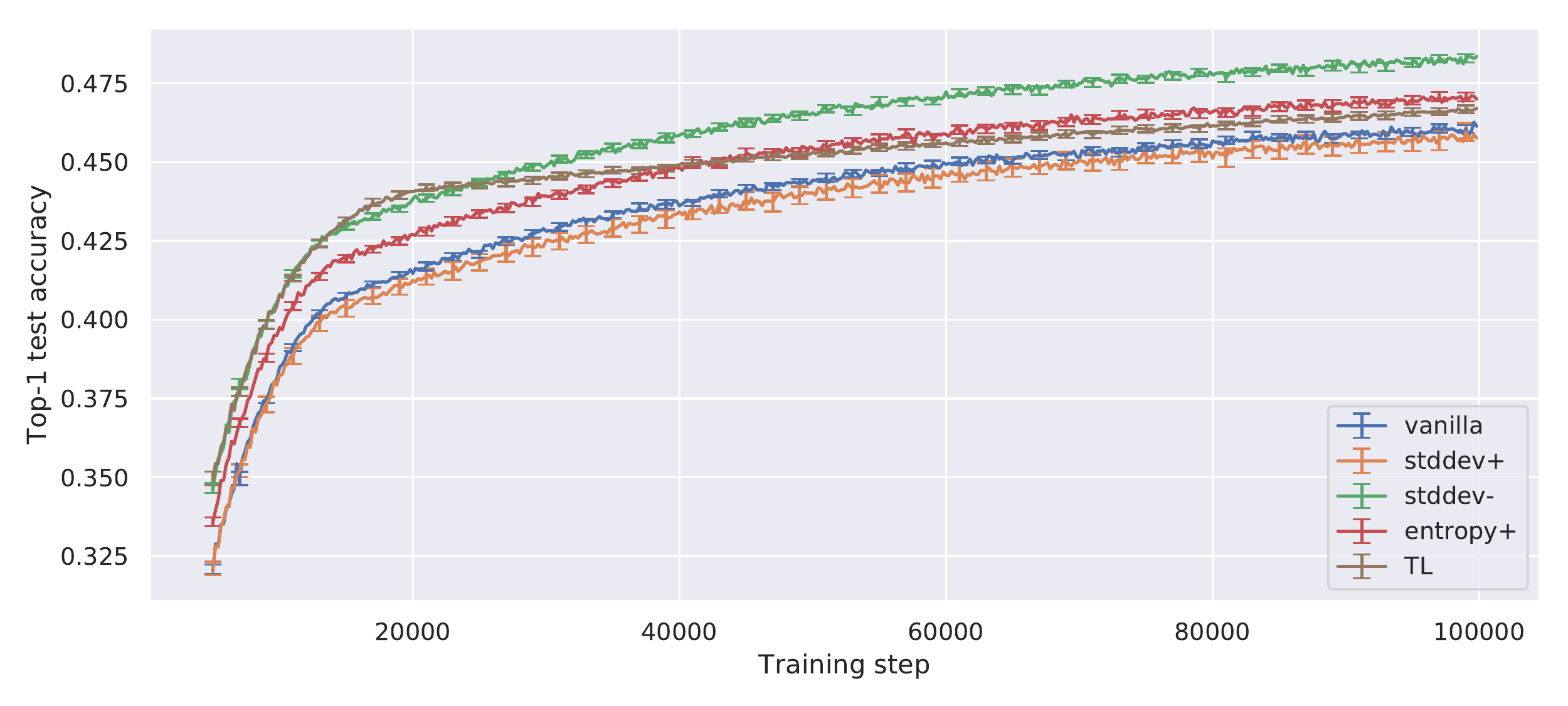} \\
  \includegraphics[width=0.8\linewidth]{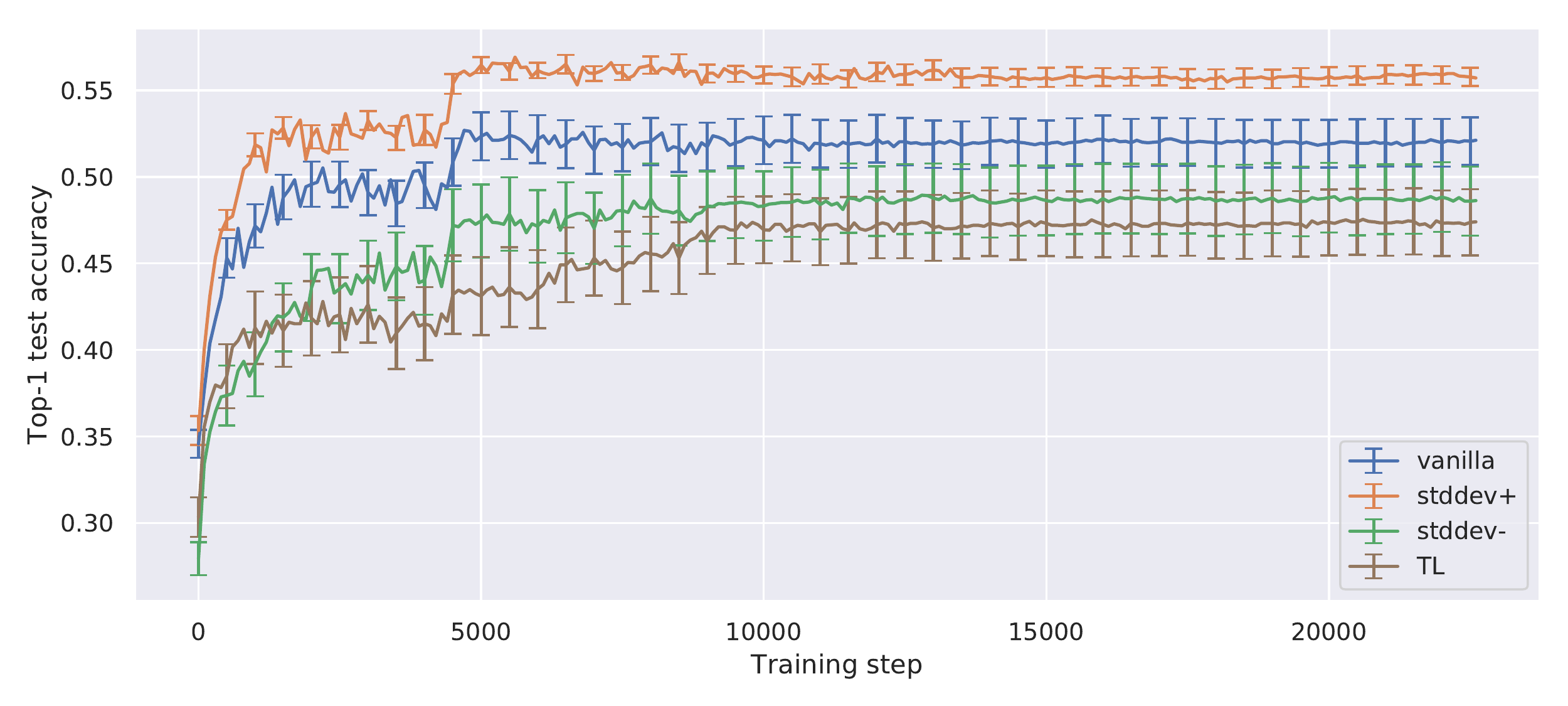}
  \vspace{-.1in}
  \caption{Learning curves for Cases $\displaystyle 3$ (top row: CNN-8 + CIFAR-100) and $\displaystyle 4$ (bottom row: CNN-8 + ImageNet Cats). Error bars represent the standard error of the mean (STE) after 25 and 10 independent trials.}
  \label{fig:stddevcurves}
  \vspace{-.1in}
\end{figure*}

\section{Statistical measures for defining curricula} 
\label{statisticalmeasuresfordefiningcurricula}

In this section, we discuss our simple approach of using statistical measures to define curricula for real image classification tasks. \citet{letsagree} shows that the orders in which a dataset is learned by various network architectures are highly correlated. While training a stronger learner, it first learns the examples learned by a weaker learner, and then continues to learn new examples. Can we design an explicit curriculum that sorts the examples according to the implicit  order in which they are learned by a network? From Figure \ref{fig:implicitcurricula} it is clear that the CIFAR-100 images learned at the beginning of training have bright backgrounds or rich color shades, while the images learned at the end of training have monotonous colors. We observe that the CIFAR-100 images learned at the beginning of training have a higher mean standard deviation ($= 0.25$) than those learned at the end of training ($ = 0.19$). For MNIST, the mean standard deviation of images learned at the beginning of training ($= 0.23$) is lesser than those learned at the end of training ($= 0.25$). Motivated by this observation, we investigate the benefits of using standard deviation for defining curriculum scoring functions in order to improve the generalization of the learner. We perform mutiple experiments and validate our proposal over various image classification datatsets with different network architectures.

Standard deviation and entropy are informative statistical measures for images and used widely in digital image processing (DIP) tasks \citep{impstatmeasure, entropy}. \citet{radarstddev} uses standard deviation filters for effective edge-preserving smoothing of radar images. Natural images might have a higher standard deviation if they have a lot of edges and/or vibrant range of colors. Edges and colours are among the most important features that help in image classification at a higher level. Figure \ref{fig:stddevimages} shows 8 images which have the lowest and highest standard deviations in the CIFAR-100 dataset. Entropy gives a measure of image information content and is used for various DIP tasks such as automatic image annotation \citep{entropyforimgannotation}.

We experiment using the standard deviation measure ($\displaystyle stddev$), the Shanon's entropy measure ($\displaystyle entropy$) \citep{shannon}, and different norm measures as scoring functions for CL (see Algorithm \ref{algo:std}). The performance improvement with norm measures is not consistent and significant  over the experiments we perform (see Suppl. \ref{app:aboutnorms} for details). For a flattened image example represented as $\displaystyle \vx_i = [x^{(0)}_i, x^{(1)}_i, ..., x^{(d-1)}_i]^\textrm{T} \in [-1, 1]^d$, we define
\begin{equation}
\begin{split}  \label{score_std}
\displaystyle
\mu(\vx_i) = \frac{\sum_{j=0}^{d-1} x^{(j)}_i}{d}
\textrm{~~~~~~~~~~~~~~~~and}\\
 stddev(\vx_i) = \sqrt{\frac{\sum_{j=0}^{d-1} (x^{(j)}_i - \mu(\vx_i ) )^2}{d}}.
\end{split}
\end{equation}
We use a fixed exponential pace function that exponentially increases the amount of data exposed to the network after every fixed $\displaystyle step\_length$ number of training steps. For a training step $\displaystyle i$, it is formally given as:
$ \displaystyle pace(i) = \floor{\min (1, starting\_fraction \cdot inc^{\floor{\frac{i}{step\_length}}}) \cdot N}$,
where $\displaystyle starting\_fraction$ is the fraction of the data that is exposed to the model initially, $\displaystyle inc$ is the exponential factor by which the the pace function value increases after a step, and $\displaystyle N$ is the total number of examples in the data.

\subsection{Baselines}
\label{statbaselines}

We use \textit{vanilla} and CL by transfer learning \citep{powerofcl}, denoted as \textit{TL}, as our baselines. We use the same hyperparameters and the codes\footnote{\url{https://github.com/GuyHacohen/curriculum_learning}} published by the authors for running \textit{TL} experiments. \textit{TL} works with the aid of an Inception network \citep{inception} pre-trained on the ImageNet dataset \citep{imagenet}. The activation levels of the penultimate layer of this Inception network is used as a feature vector for each of the images in the training data. These features are used to train a classifier (e.g, support vector machine) and its confidence scores for each of the training images are used as the curriculum scores.

\subsection{Experiments}
\label{statexpsetup}

We denote CL models with scoring functions $\displaystyle stddev$ as \textit{stddev+}, $\displaystyle -stddev$ as \textit{stddev-}, $\displaystyle entropy$ as \textit{entropy+}, and
$\displaystyle -entropy$ as \textit{entropy-}. Our codes are published on GitHub\footnote{\url{https://github.com/vinusankars/curriculum_learning}}. We employ three network architectures for our experiments: a) FCN-$512$ -- A $\displaystyle 2$-layer fully-connected network (FCN-$m$) with $\displaystyle m = 512$ hidden neurons with Exponential Linear Unit (ELU) nonlinearities, b) CNN-$8$ \citep{powerofcl} -- A moderately deep CNN with $\displaystyle 8$ convolution layers and $\displaystyle 2$ fully-connected layers, and c) ResNet-20 \citep{resnet} -- A deep CNN.

We use the following datasets for our experiments: a) MNIST, b) Fashion-MNIST, c) CIFAR-10, d) CIFAR-100, e) Small Mammals (a super-class of CIFAR-100, \citep{mammals}), and f) ImageNet Cats (a subset of 7 classes of cats in ImageNet, see Suppl. \ref{app:dataset}). For our experiments, we use the same setup as used in \citet{powerofcl}. We use learning rates with an exponential step-decay rate for the optimizers in all our experiments as traditionally done \citep{practise1, practise2}. In all our experiments, the models use fine-tuned hyperparameters for the purpose of an unbiased comparison of model generalization over the test set. More experimental details are deferred to Suppl. \ref{app:expdetails}. While practically performing model training, we prioritize class balance. Although we do not follow the exact ordering provided by the curriculum scoring function, the ordering within a class is preserved.

We define 9 test cases. Cases 1--5 use CNN-$8$ to classify  Small Mammals, CIFAR-10, CIFAR-100, ImageNet Cats, and Fashion-MNIST datasets, respectively. Cases 6--8 use FCN-$512$ to classify  MNIST, Fashion-MNIST, and CIFAR-10 datasets, respectively. Case 9 uses ResNet-20 to classify the ImageNet Cats dataset.

 \begin{figure}[t]
 \centering
    \includegraphics[width=0.45\textwidth]{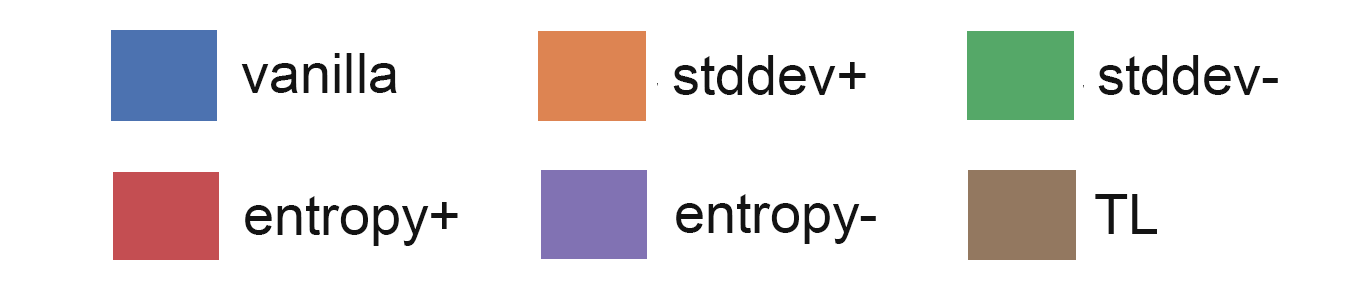}\label{fig:barlegend}
     \subfigure[Case 1]{\includegraphics[width=0.15\textwidth]{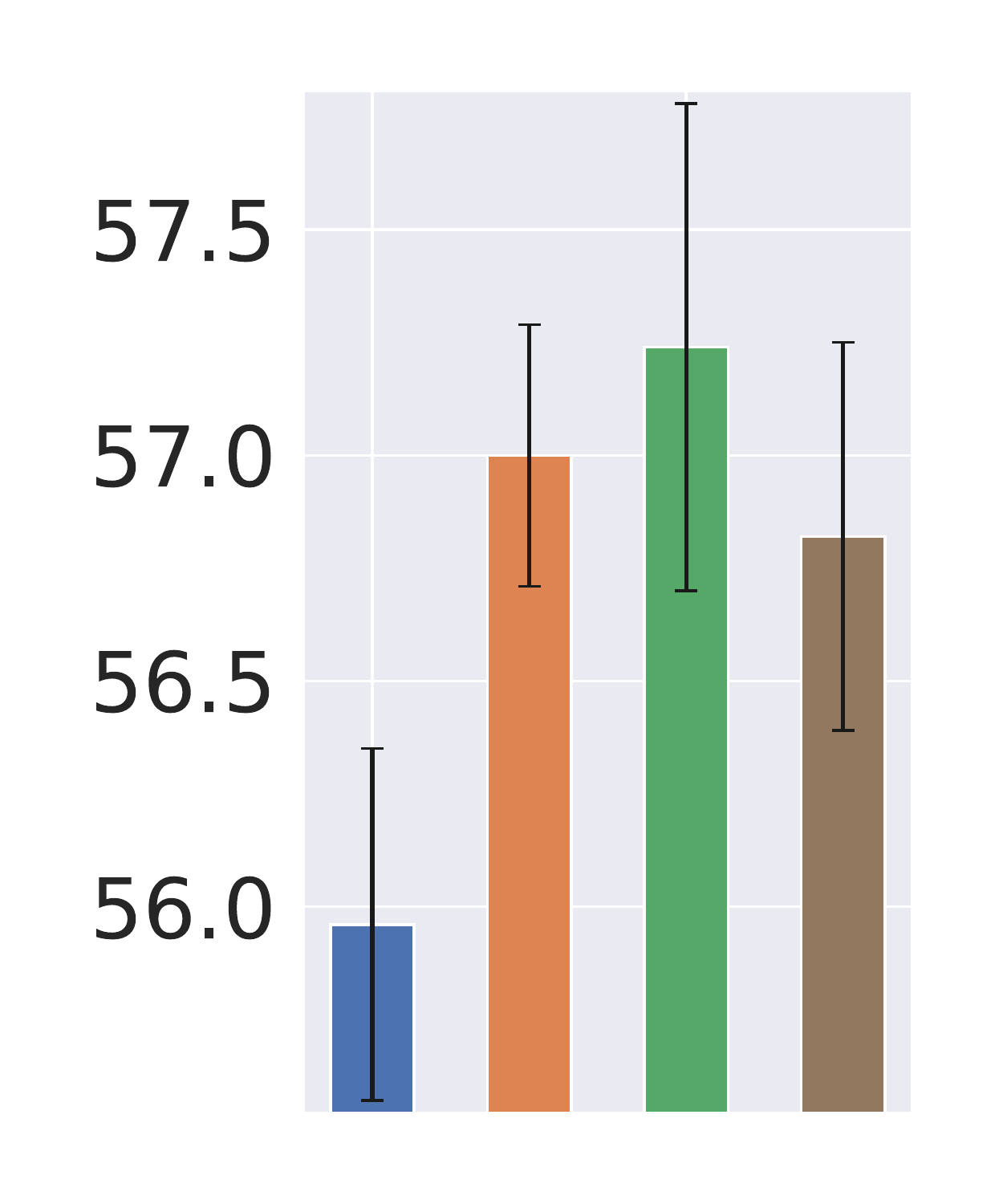}\label{fig:bar1}}
     \subfigure[Case 2]{\includegraphics[width=0.15\textwidth]{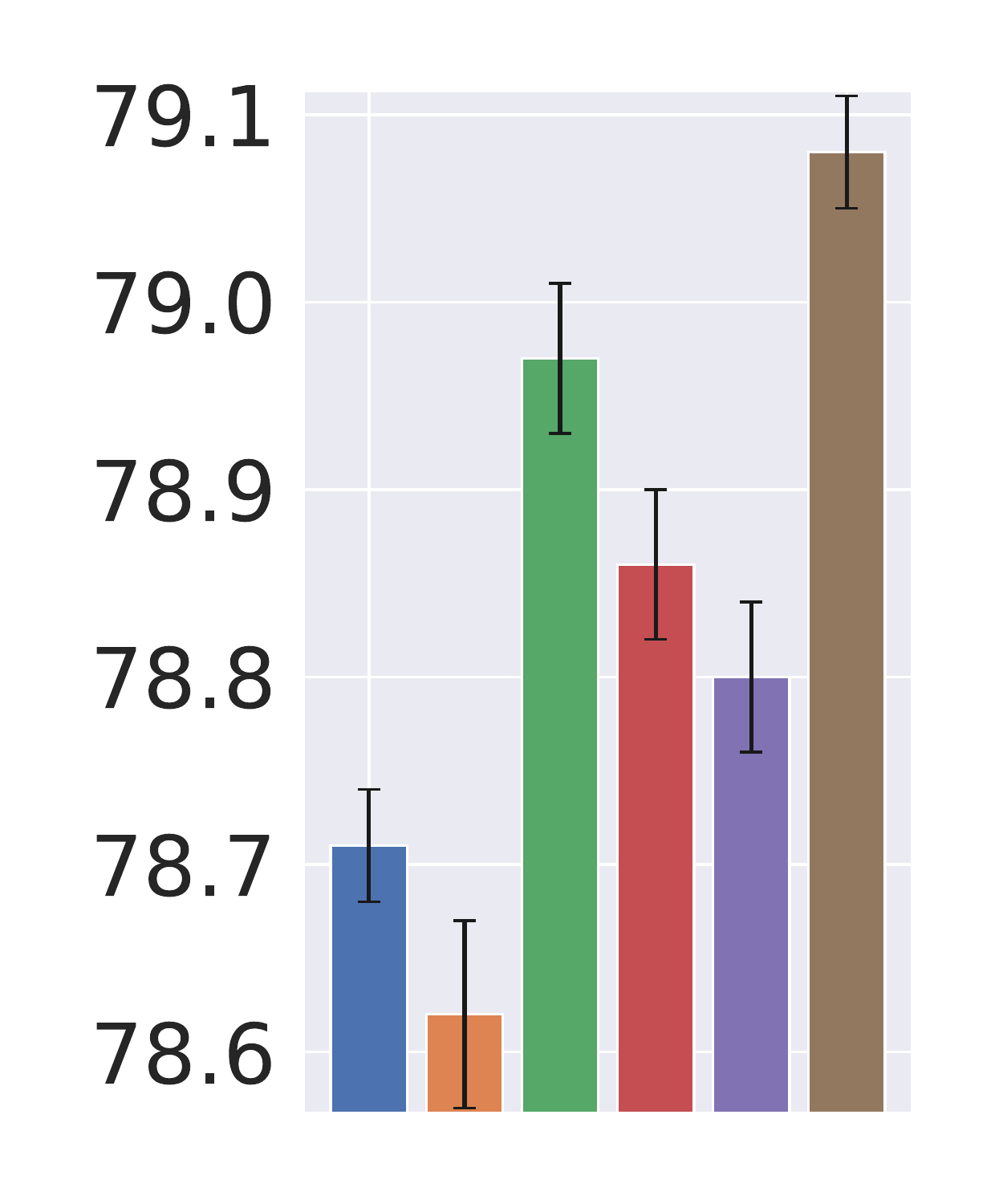}\label{fig:bar2}}
     \subfigure[Case 3]{\includegraphics[width=0.15\textwidth]{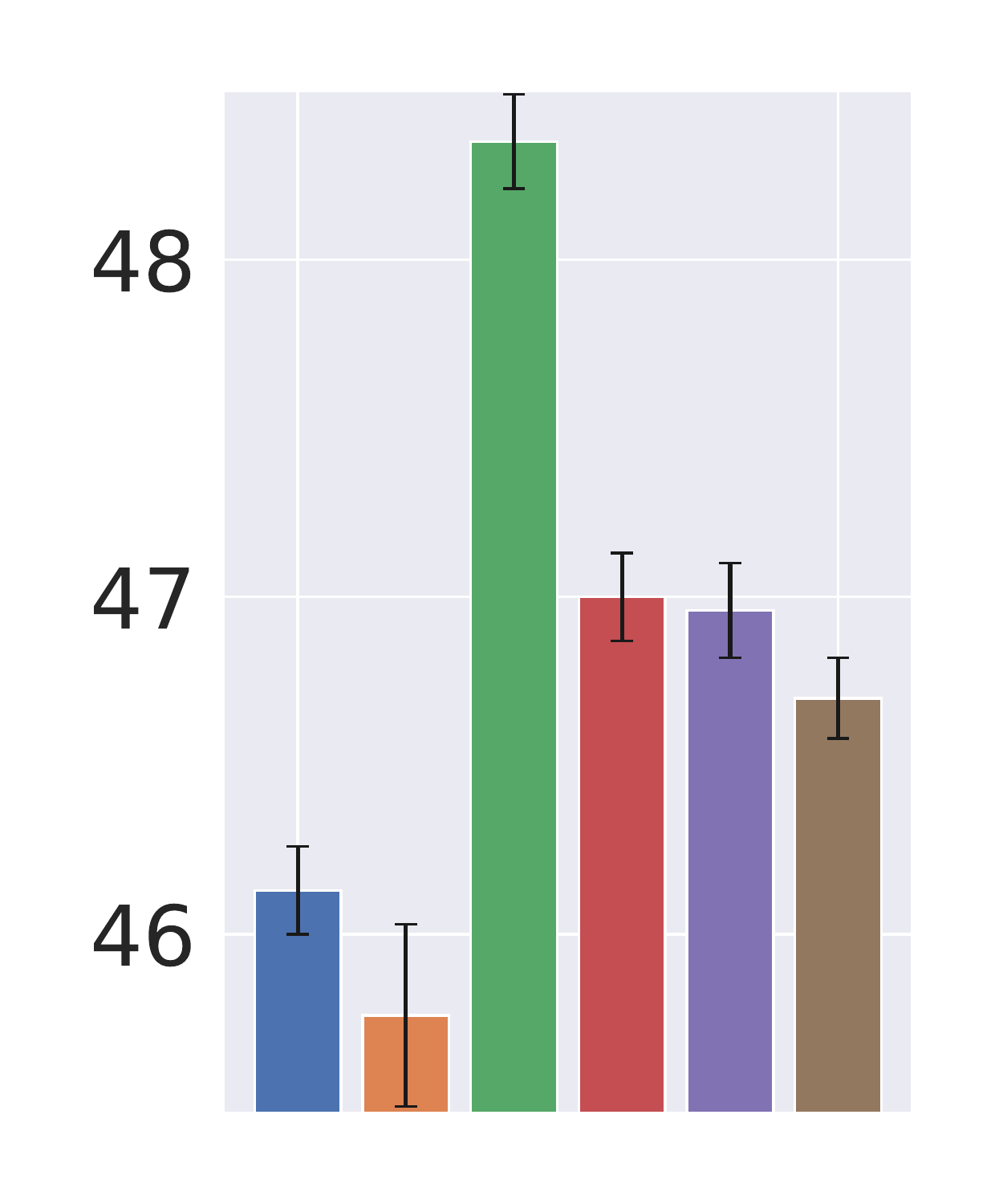}\label{fig:bar3}}
     \subfigure[Case 4]{\includegraphics[width=0.15\textwidth]{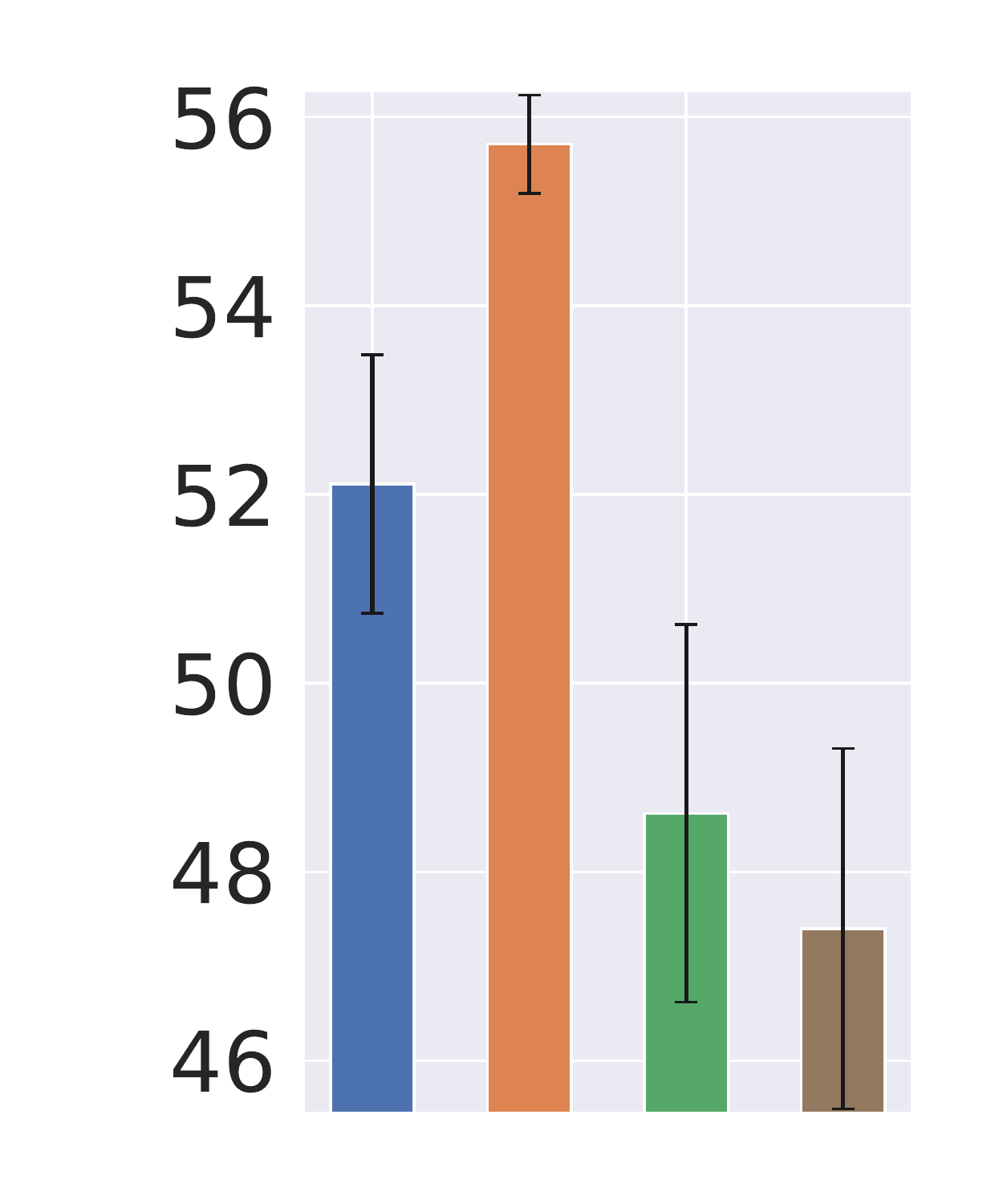}\label{fig:bar4}}
     \subfigure[Case 5]{\includegraphics[width=0.15\textwidth]{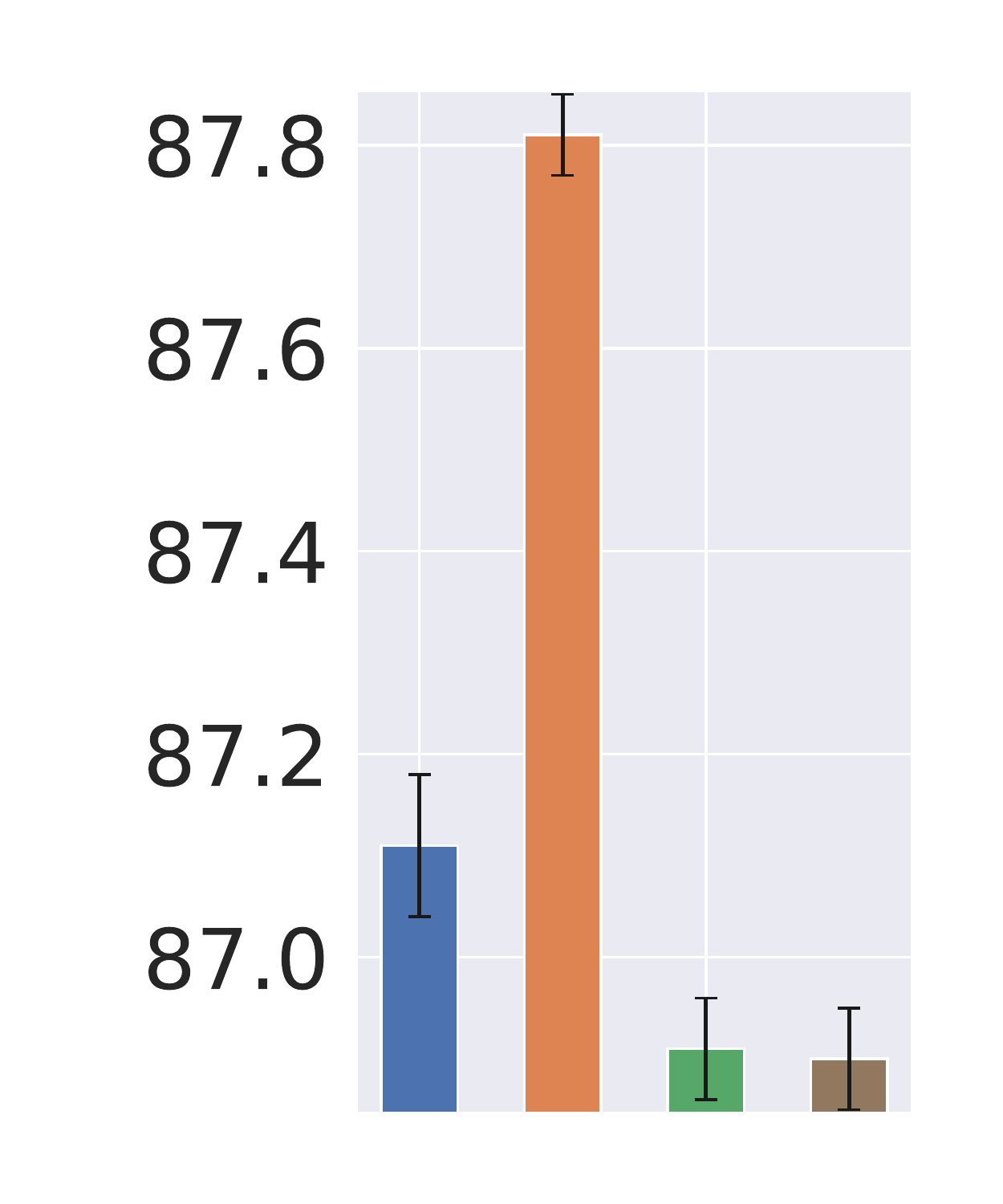}\label{fig:bar5}}
     \subfigure[Case 6]{\includegraphics[width=0.15\textwidth]{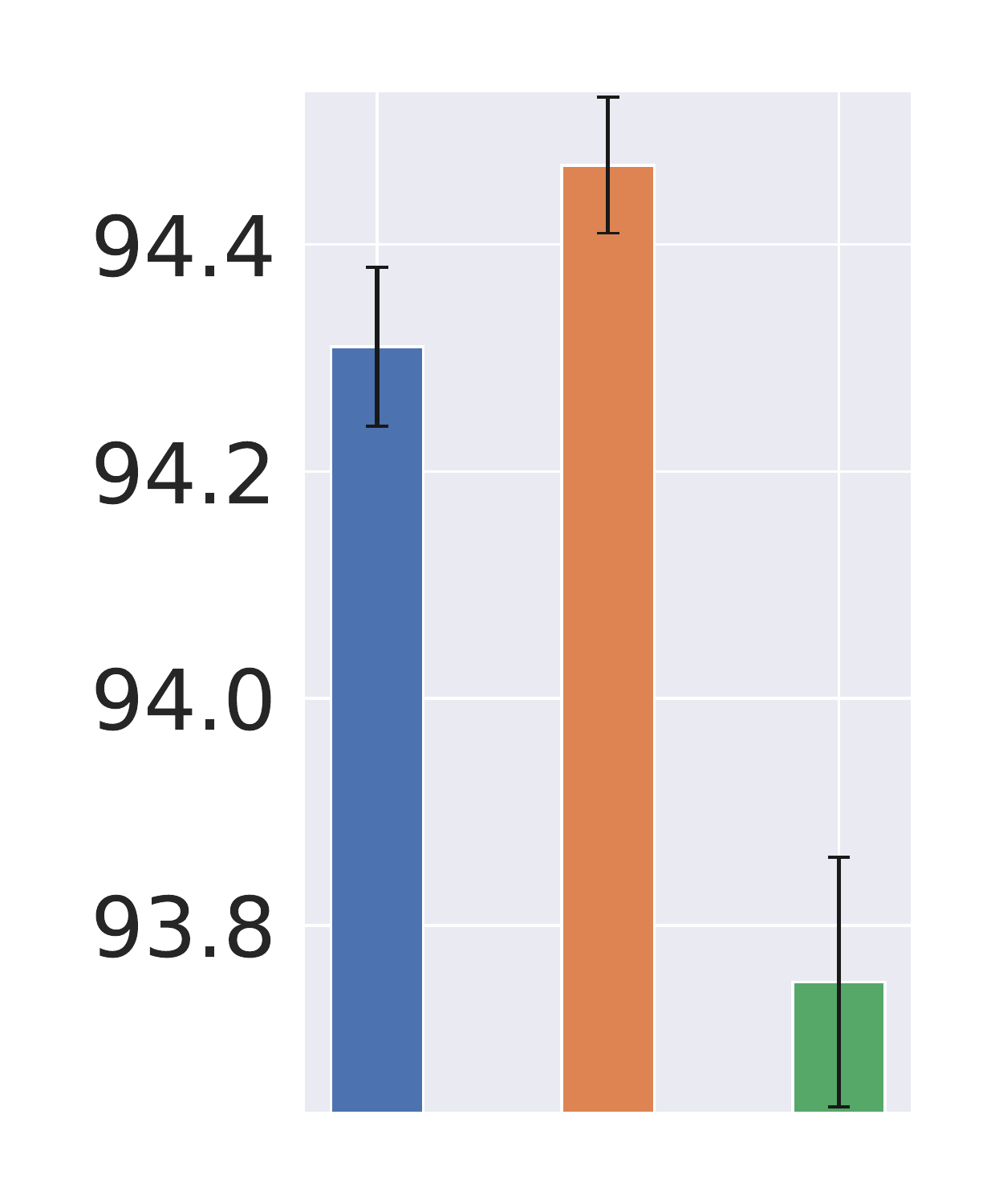}\label{fig:bar6}}
     \subfigure[Case 7]{\includegraphics[width=0.15\textwidth]{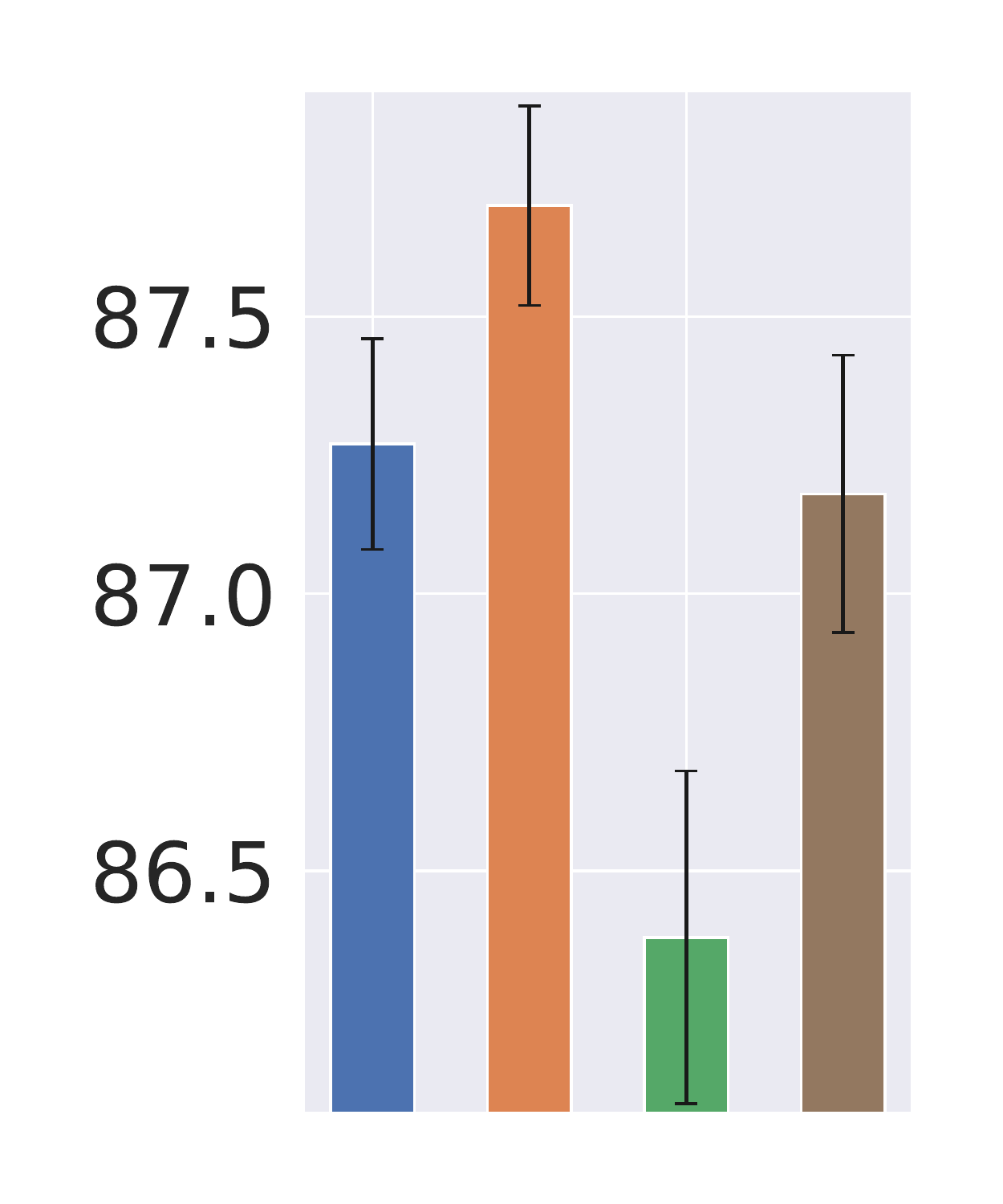}\label{fig:bar7}}
     \subfigure[Case 8]{\includegraphics[width=0.15\textwidth]{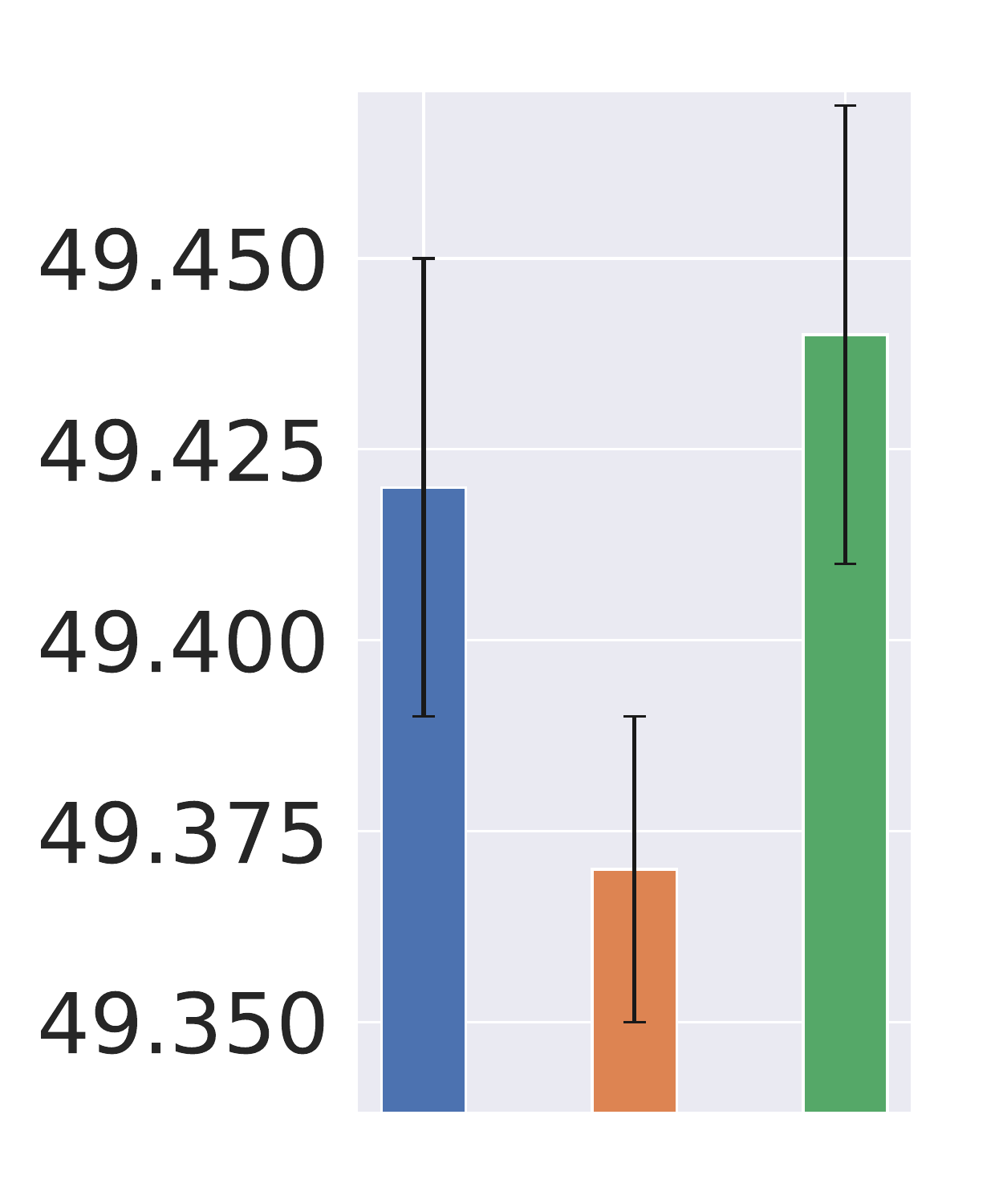}\label{fig:bar8}}
     \subfigure[Case 9]{\includegraphics[width=0.15\textwidth]{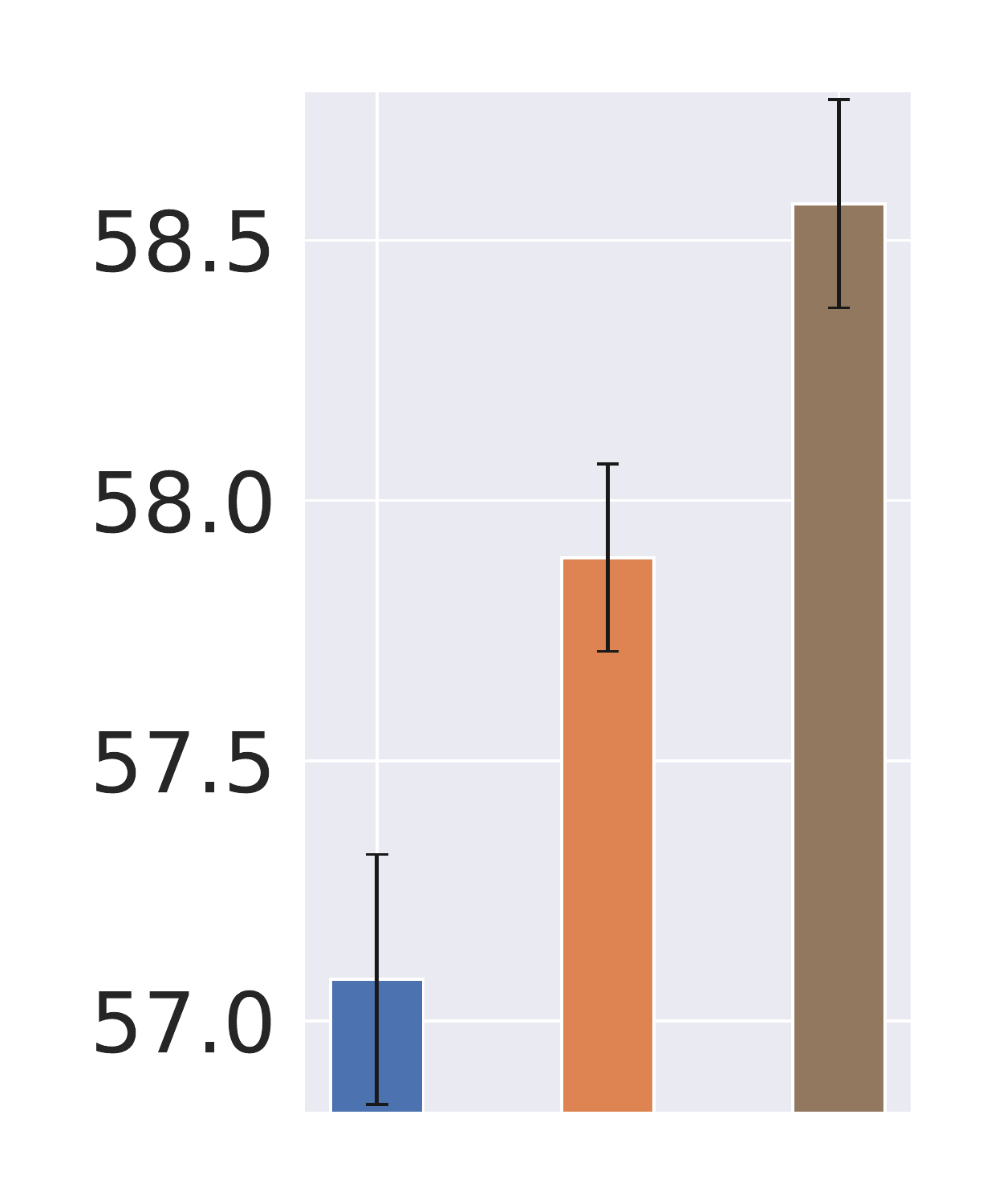}\label{fig:bar9}}

     \caption{Bars represent the final mean top-1 test accuracy (in $\displaystyle \%$) achieved by models in Cases 1--9. Error bars represent the STE after 25 independent trials for Cases 2, 3, 5 -- 8, and 10 independent trials for Cases 1, 4, 9.}
     \label{fig:bars}
     \vspace{-.2in}
     
 \end{figure}
 
Figure \ref{fig:stddevcurves} shows the improvement in network generalization of CNN-$8$ on CIFAR-100 and ImageNet Cats datasets using \textit{stddev} CL algorithms. Figure \ref{fig:bars} shows the results of all the test cases that we perform. From Figures \ref{fig:bar2} and \ref{fig:bar3} it is clear that $stddev$ serves as a better scoring function than $entropy$. Further, we  observe that the datasets MNIST, Fashion-MNIST, and ImageNet Cats best follow the curriculum variant \textit{stddev+}. CIFAR-100, CIFAR-10, and Small Mammals follow the curriculum defined by \textit{stddev-}. As discussed earlier in this section, this trend is consistent with the \textit{stddev} order in which the dataset images are implicitly learned by the network.  In all the test cases, \textit{stddev} CL algorithm consistently performs better than \textit{vanilla} with a mean improvement of $\sim 1.05\%$ top-1 test accuracy.

\begin{table}[t]
\caption{\textit{stddev} curriculum selection using median pixel distance values. Bolded value corresponds to the curriculum variant that works the best for a dataset.}
\label{tab:medians}
\vskip 0.15in
\begin{center}
\begin{small}
\begin{sc}
\begin{tabular}{ccc}
\toprule
Dataset & $M_+$ & $M_-$ \\
\midrule
MNIST   & \textbf{0.01}  & 0.00 \\
Fashion-MNIST   & \textbf{0.06}  & 0.01 \\
Small Mammals   & 0.03  & \textbf{0.08} \\
CIFAR-10   & 0.02  & \textbf{0.06} \\
CIFAR-100   & 0.06  & \textbf{0.09} \\
ImageNet Cats   & \textbf{0.07}  & 0.02 \\
\bottomrule
\end{tabular}
\end{sc}
\end{small}
\end{center}
\vskip -0.1in
\end{table}

Let $med(\vx)$\footnote{Similar to \href{https://github.com/numpy/numpy/blob/v1.20.0/numpy/lib/function_base.py}{NumPy median} function.} denote the median value of all the pixels in image(s) $\vx$ and $M = med([\vx_i]_{i=0}^{N-1})$ denote the median pixel value of the full training images, where the examples are ordered according to \textit{stddev+}. We denote $M_+ = \lvert M - med([\vx_i]_{i=0}^{b-1}) \rvert$ and $M_- = \lvert M - med([\vx_i]_{i=N-b}^{N-1}) \rvert$ as the median pixel distances of the first batches of examples sampled according to \textit{stddev+} and \textit{stddev-}, respectively, where $b$ is the batch size.  Interestingly, we notice that the \textit{stddev} curriculum variant that best works for a dataset has a higher median pixel distance as shown in Table \ref{tab:medians}.

 \begin{figure}[b]
 \centering
    \includegraphics[width=0.4\textwidth]{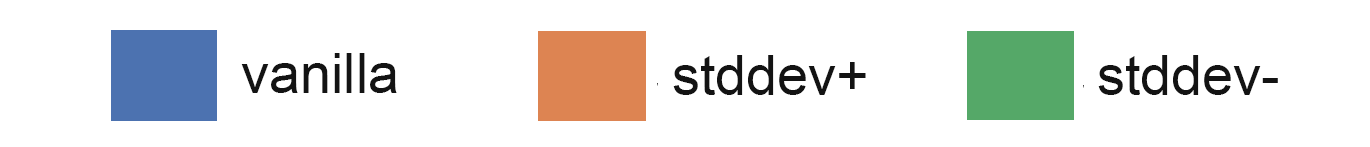}\label{fig:noisebarlegend}
     \subfigure[CIFAR-100]{\includegraphics[width=0.15\textwidth]{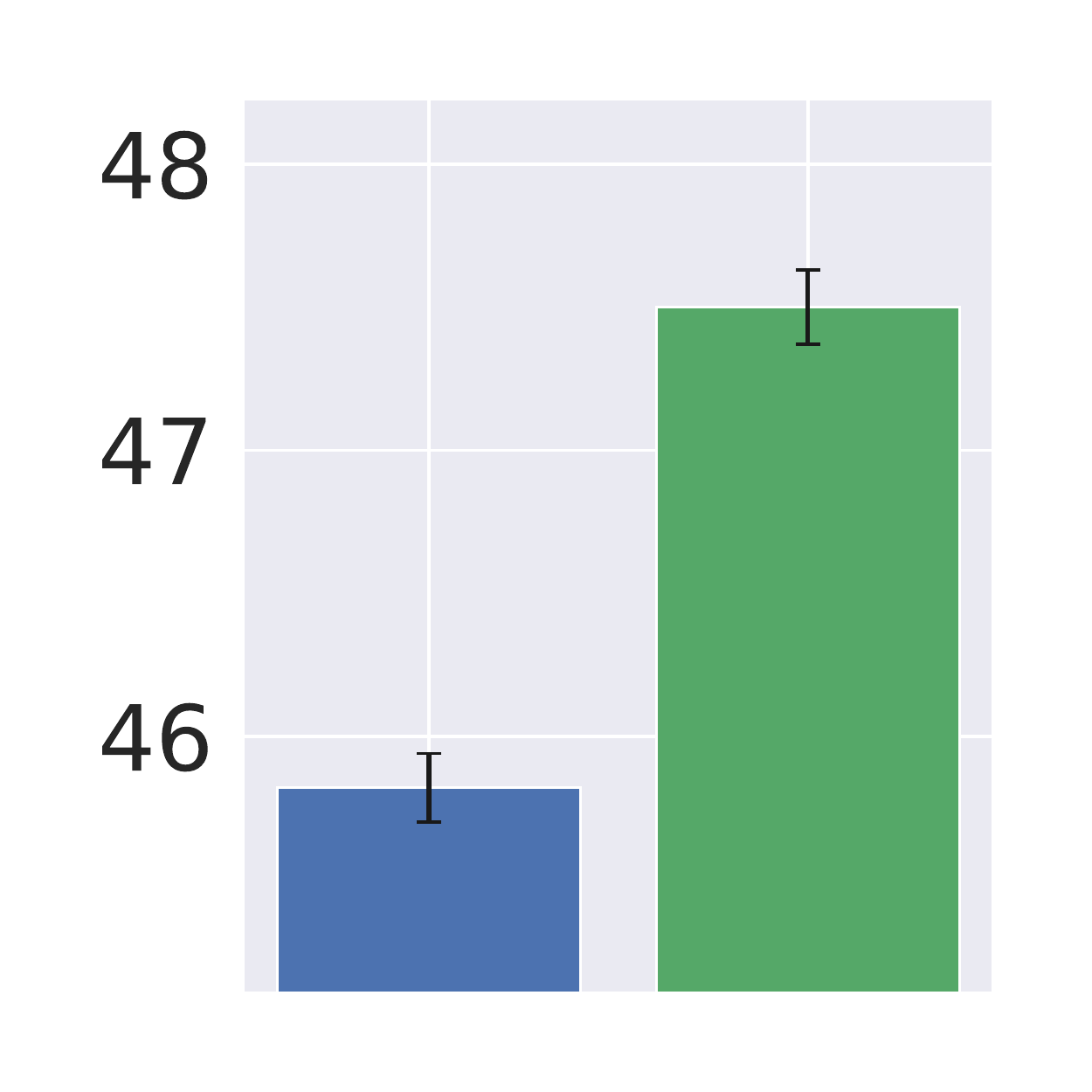}\label{fig:noisebar10}} 
     \subfigure[ImageNet Cats]{\includegraphics[width=0.15\textwidth]{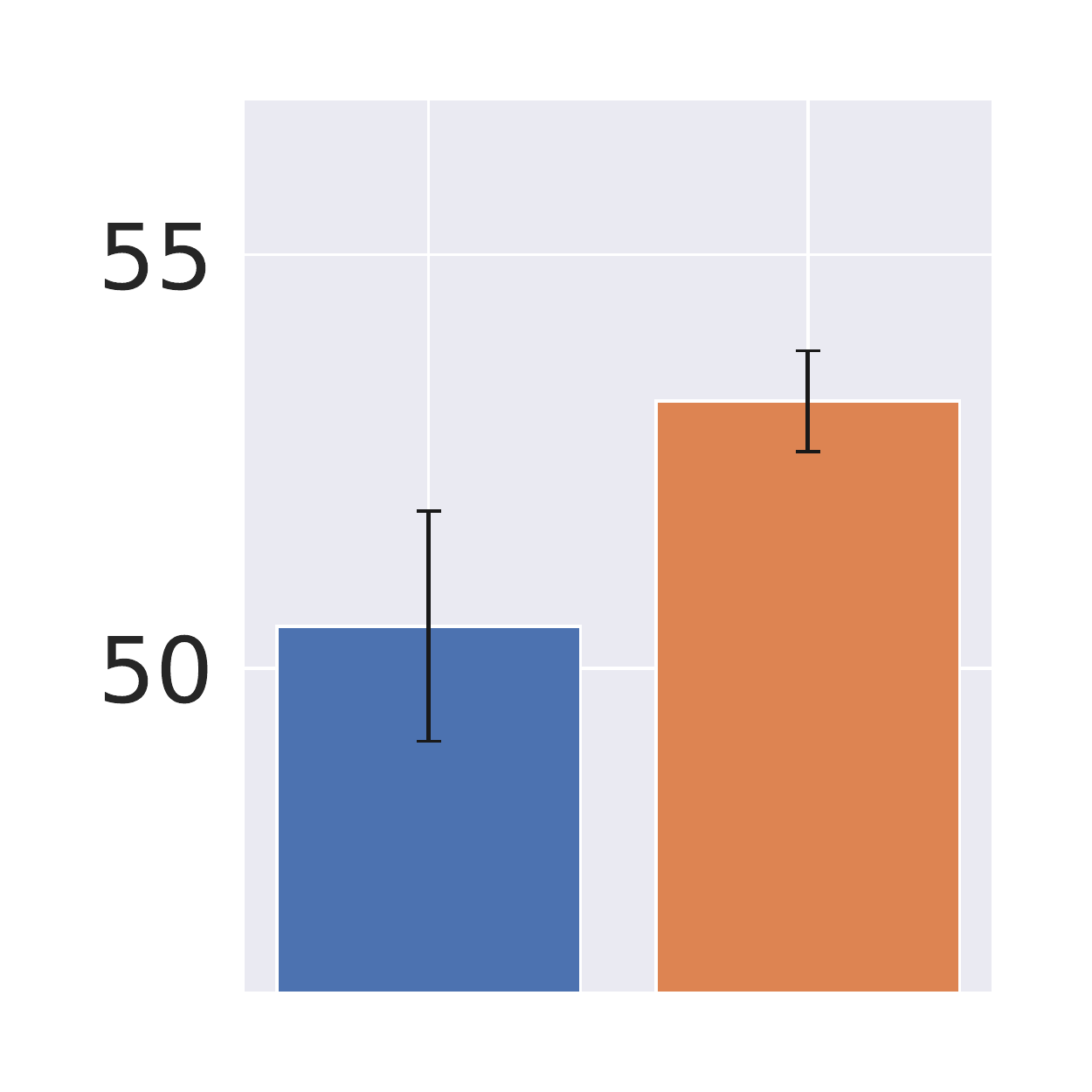}\label{fig:noisebar11}}
     \caption{Bars represent the final mean top-1 test accuracy (in $\displaystyle \%$) achieved by CNN-$8$. Error bars represent the STE after 25 and 10 independent trials, respectively.}
     \label{fig:noisebars}
     \vspace{-.1in}
     
 \end{figure}

We also test the robustness of our \textit{stddev} algorithms to noisy labels. For this purpose, we design two test cases: CNN-$8$ to classify CIFAR-100 and ImageNet Cats datasets with $20\%$ label noise. We add label noise by uniformly sampling $20\%$ of the data points and randomly changing their labels. Figure \ref{fig:noisebars} shows that our CL algorithms work well in training settings with label noise, even with only coarse fine-tuning of curriculum hyperparameters.

\begin{figure}[t]
    \centering
    \includegraphics[width=1\linewidth]{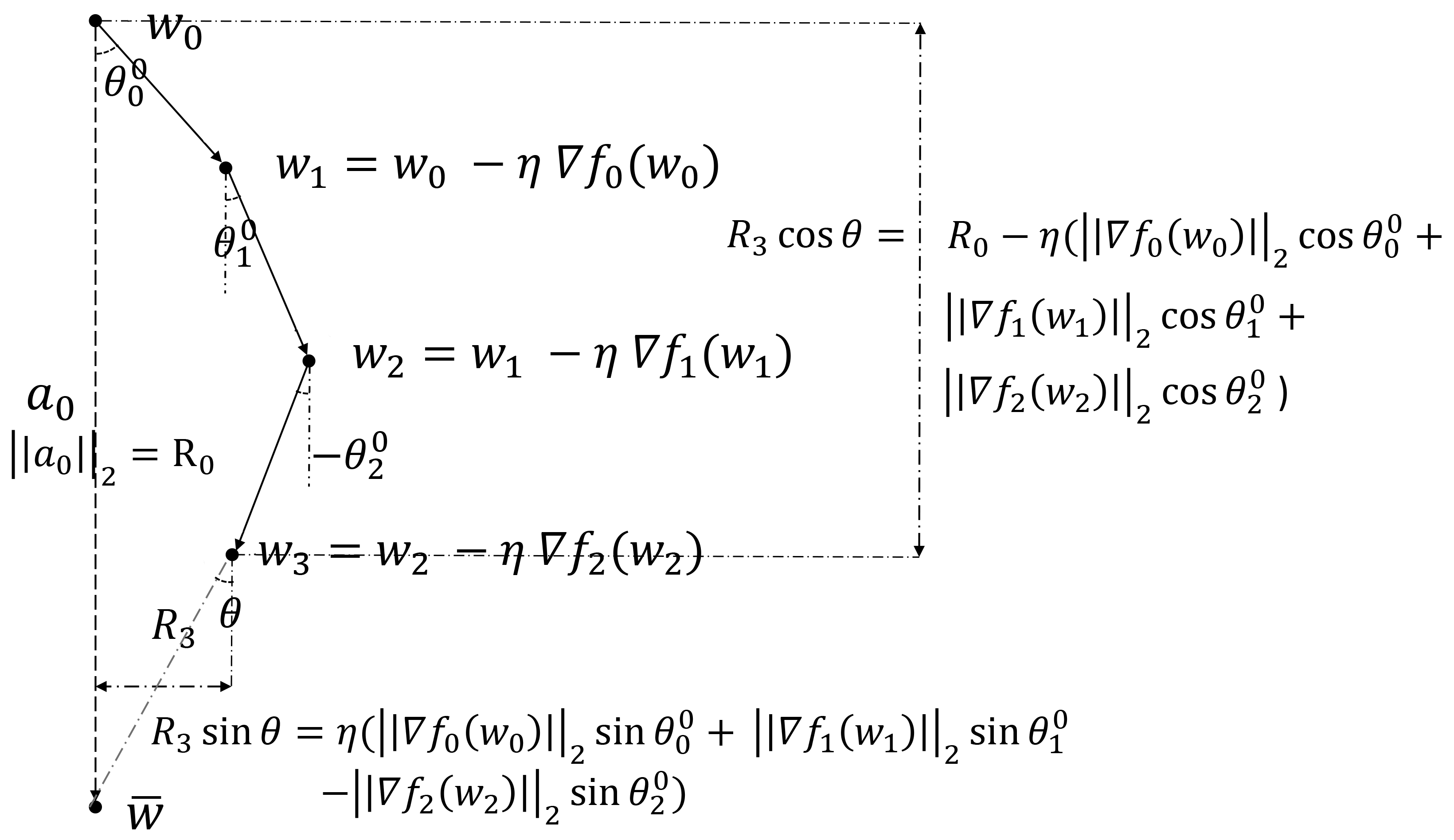}
    \caption{A geometrical interpretation of gradient steps for the understanding of equation \ref{eqn:rt2}.}
    \label{fig:geometry}
\end{figure}

 \begin{algorithm}[t]
   \caption{Dynamic curriculum learning (\textit{DCL+}).}
   \label{algo:dcl}
\begin{algorithmic}
   \STATE {\bfseries Input:} Data $\displaystyle \mathcal{X}$, local minima $\displaystyle \bar{\vw}$, weight $\displaystyle \vw_t$, batch size $\displaystyle b$, and pacing function $\displaystyle pace$. 
   \STATE {\bfseries Output:} Sequence of mini-batches $\displaystyle B_t$ for the next training epoch.
   \STATE $\displaystyle {\va}_t \gets \bar{\vw} - \vw_t$ \STATE $\displaystyle \rho_t \gets [~]$ 
   \STATE $\displaystyle B_t \gets [~]$
   \FOR{$i=0$ {\bfseries to} $N-1$}
    \STATE $\displaystyle \rho_{t,i} \gets \displaystyle-\frac{ {\va}_t^\textrm{T} \cdot\nabla f_i(\vw_t)}{\|{\va}_t\|_2}$.
    \ENDFOR
    \STATE $\displaystyle {\mathcal{X}_t} \gets \mathcal{X}$ sorted according to $\displaystyle \rho_{t,i}$, in ascending order
    \STATE $\displaystyle size \gets pace(t)$
    \FOR{$\displaystyle (i = 0;~size;~b)$}
    \STATE append $\displaystyle {\mathcal{X}_t}[i,...,i+b-1]$ to $B_t$
    \ENDFOR
   \STATE {\bfseries return} $\displaystyle B_t$
\end{algorithmic}
\end{algorithm}

\section{Dynamic Curriculum Learning} 
\label{section:dcl}

For DCL algorithms \citep{spl}, examples are either scored and sorted or automatically selected \citep{autocl, tscl} after every few training steps since the scoring function changes dynamically with the learner as training proceeds. \citet{powerofcl} and \citet{cl} use a fixed scoring function and pace function for the entire training process. They empirically show that a curriculum helps to learn fast in the initial phase of the training process. In this section, we propose our novel DCL algorithm for studying the behaviour of CL. Our DCL algorithm updates the difficulty scores of all the examples in the training data at every epoch using their gradient information. 

We hypothesize the following: Given a weight initialization $\vw_0$ and a local minima $\bar{\vw}$ obtained by full training of \textit{vanilla} SGD, the curriculum ordering determined by our DCL variant leads to convergence in fewer number of training steps than \textit{vanilla}. We first describe the algorithm, then the underlying intuition,  and finally validate the hypothesis using experiments. 

Our DCL algorithm iteratively works on reducing the L$2$ distance, $\displaystyle R_t$, between the weight parameters $\displaystyle {\vw}_t$ and $\displaystyle \bar{\vw}$ at any training step $\displaystyle t$. Suppose, $S_t$ is the index of the example sampled at training step $t$, and for any $\tilde{t} < \displaystyle t$, $\displaystyle S_{\tilde{t},t}$  is the ordered set containing the $\displaystyle (t - \tilde{t} + 1)$ indices of training examples that are to be shown to the learner from the training steps $\displaystyle \tilde{t}$ through $\displaystyle t$. Let us define $\displaystyle {\va}_t = (\bar{\vw} - {\vw}_t)$, $\displaystyle R_t = {\| {\va}_t \|}_2$, and $\displaystyle {\theta}_{t}^{\tilde{t}}$ as the angle between $\displaystyle \nabla f_{S_t}({\vw}_t)$ and $\displaystyle {\va}_{\tilde{t}}$. Then, using a geometrical argument, (see Figure \ref{fig:geometry}),

\begin{align} \label{eqn:rt2}
\displaystyle
R_t^2 &= \bigg(R_{\tilde{t}} - \eta ~\sum_{j={\tilde{t}}}^{j=t-1} \Big(\|\nabla f_{S_j}(\vw_j)\|_2 ~\textrm{cos} ~\theta_j^{\tilde{t}} \Big)\bigg)^2 \nonumber\\ 
&~~~~+ \eta^2 ~\bigg(\sum_{j={\tilde{t}}}^{j=t-1} \Big(\| \nabla f_{S_j}(\vw_j)\|_2 ~\textrm{sin} ~\theta_j^{{\tilde{t}}}\Big)\bigg)^2 \nonumber\\
&= R_{\tilde{t}}^2 - 2\eta R_{\tilde{t}} ~\sum_{j={\tilde{t}}}^{j=t-1} \Big(\| \nabla f_{S_j}(\vw_j) \|_2 ~\textrm{cos} ~\theta_j^{{\tilde{t}}} \Big) \nonumber\\&~~~~+ \eta^2 ~\bigg(\sum_{j={\tilde{t}}}^{j=t-1} \Big(\| \nabla f_{S_j}(\vw_j)\|_2 ~\textrm{cos} ~\theta_j^{{\tilde{t}}}\Big)\bigg)^2\nonumber \\
&~~~~+ \eta^2~\bigg(\sum_{j={\tilde{t}}}^{j=t-1} \Big(\| \nabla f_{S_j}(\vw_j)\|_2 ~\textrm{sin} ~\theta_j^{{\tilde{t}}}\Big)\bigg)^2
\end{align}

For a \textit{vanilla} model, $\displaystyle S_{0, T}$ is generated by uniformly sampling indices from $\displaystyle [N]$ with replacement. Since, finding an ordered set $\displaystyle S_{0, T}$ to minimize $\displaystyle R_T^2$ is computationally expensive, we approximate the DCL algorithm (\textit{DCL+}, see Algorithm \ref{algo:dcl}) by neglecting the terms with coefficient $\displaystyle \eta^2$ in equation \ref{eqn:rt2}.
 Algorithm \ref{algo:dcl} uses a greedy approach to  approximately minimize $\displaystyle R_t^2$ by sampling examples at every epoch using the scoring function 

\begin{equation}
\displaystyle
\begin{split} \label{score_dcl}
score_{\tilde{t}}({\vx}_{S_{t}}) &= -{\| \nabla f_{S_{t}}({\vw}_{t}) \|}_{2}~ \textrm{cos} ~\theta_{t}^{\tilde{t}} \\&= - \frac{{\va}_{\tilde{t}}^{\textrm{T}} \cdot \nabla f_{S_{t}}({\vw}_{t})}{{\|{\va}_{\tilde{t}}\|}_{2}} = \rho_{\tilde{t}, S_{t}}.
\end{split}
\end{equation}

Let us denote the models that use the natural ordering of mini-batches greedily generated by Algorithm \ref{algo:dcl} as \textit{DCL+}. \textit{DCL-} uses the same sequence of mini-batches that \textit{DCL+} exposes to the network at any given epoch, but the order is reversed.  We empirically show that \textit{DCL+} achieves a faster and better convergence with various initializations of $\displaystyle \vw_0$.

\subsection{Experiments}
\label{dclexperimentalsetup}

 \begin{figure}[t]
 \centering
     \subfigure[Experiment 1]{\includegraphics[width=0.45\textwidth]{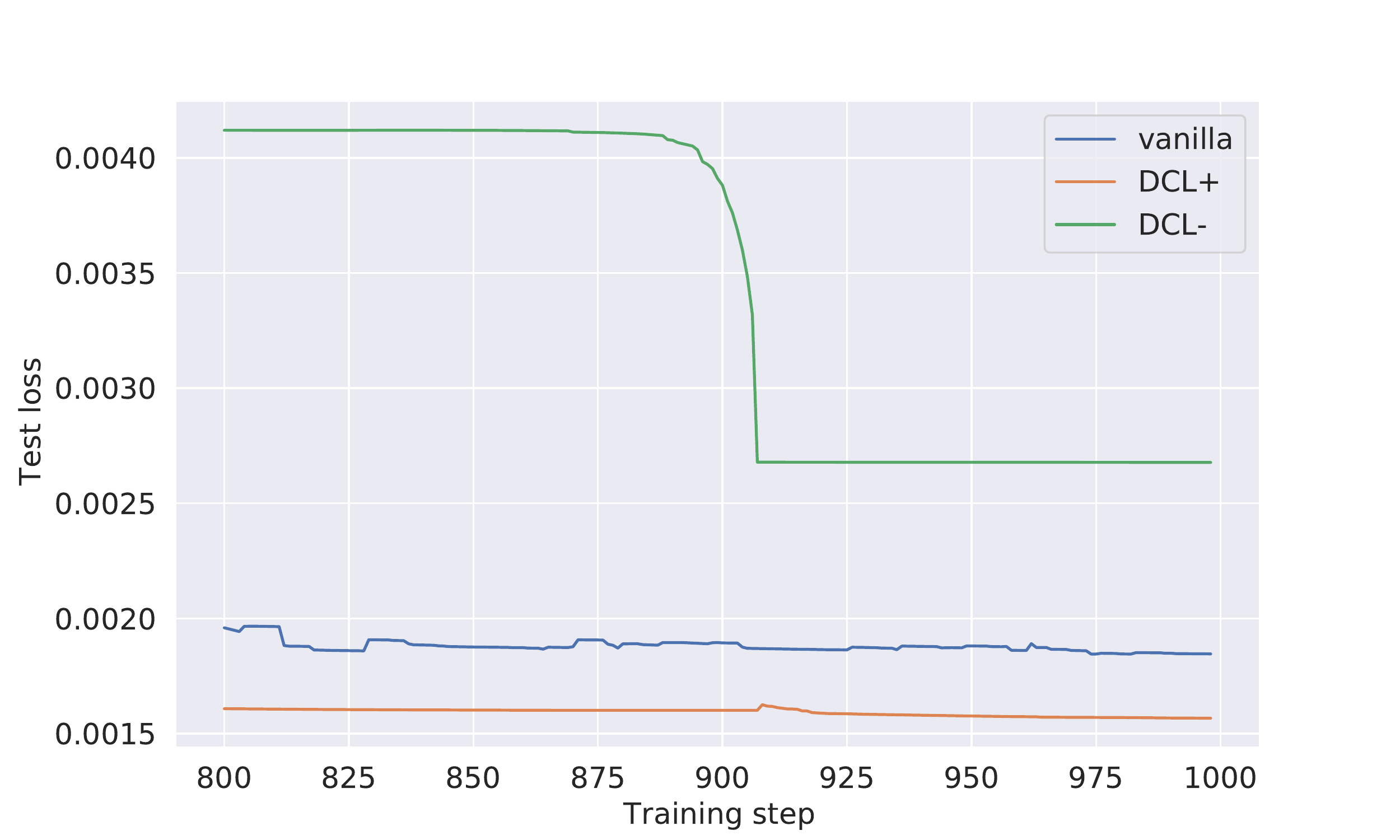}\label{fig:exp1}} 
     \subfigure[Experiment 2]{\includegraphics[width=0.45\textwidth]{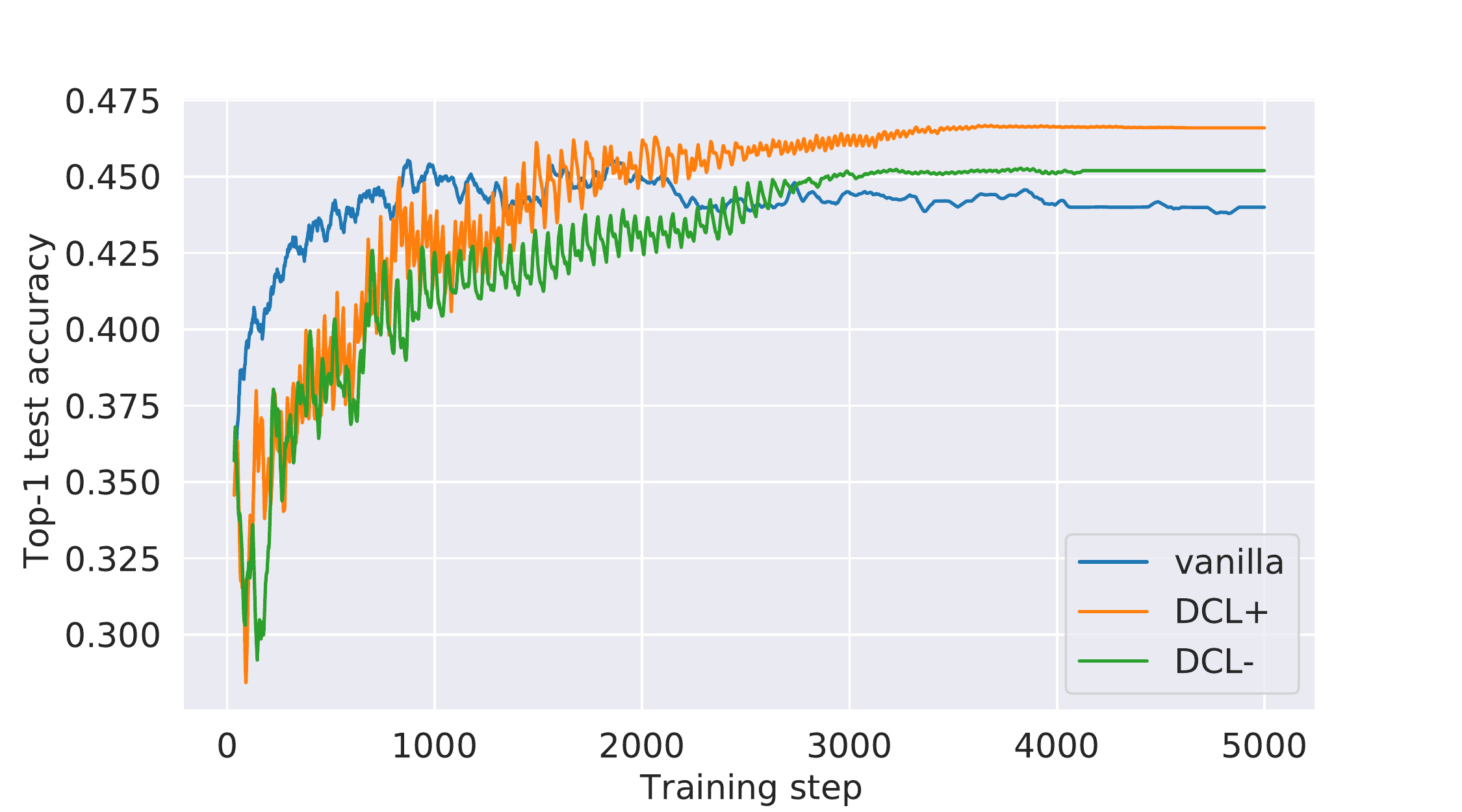}\label{fig:exp2}}
     \caption{Learning curves of experiments comparing \textit{DCL+}, \textit{DCL-}, and \textit{vanilla} SGDs.}
     \label{fig:dcl}
     \vspace{-.1in}
     
 \end{figure}
 
  \begin{figure}[t]
 \centering
 \includegraphics[width=0.45\textwidth]{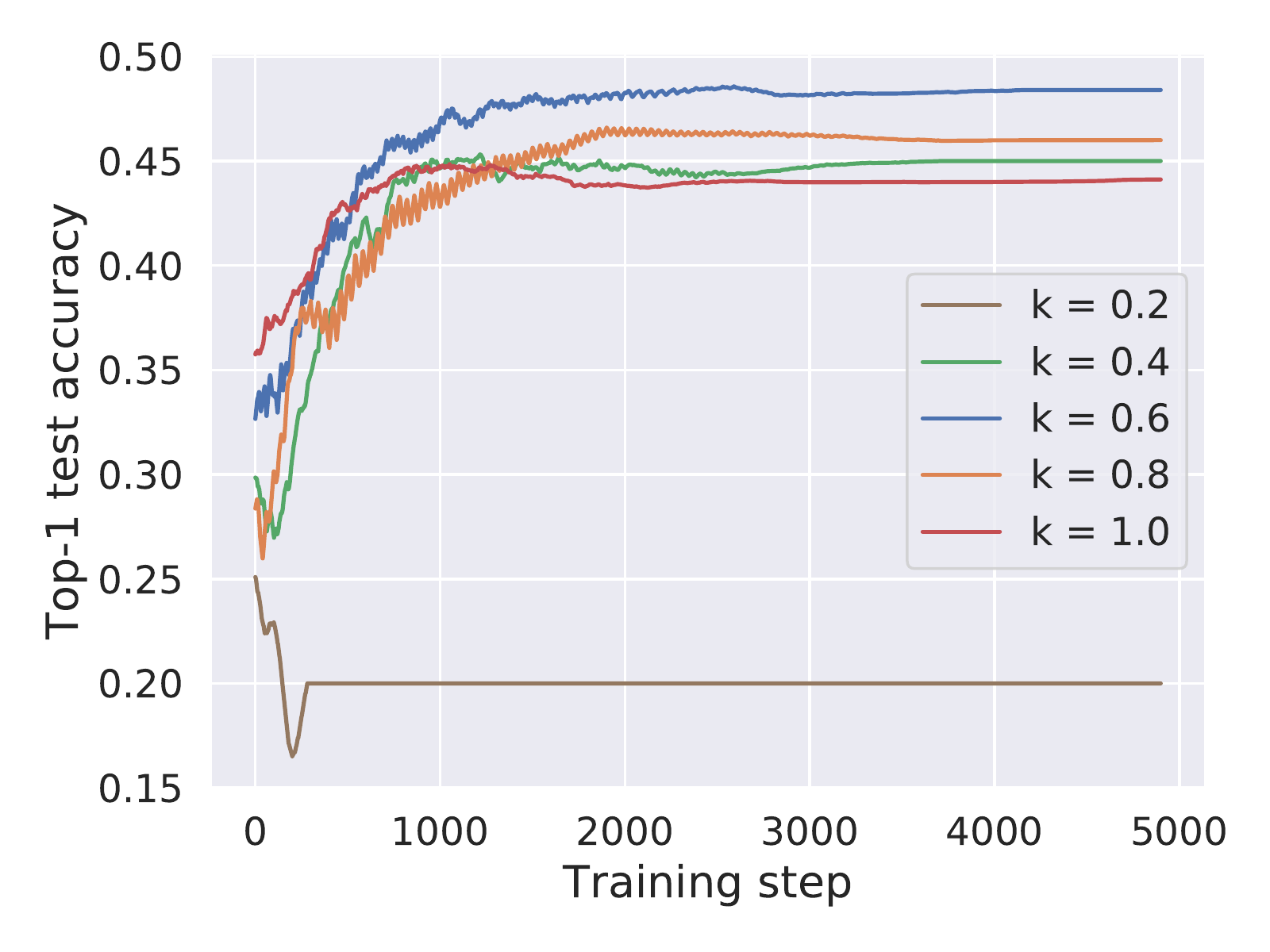}
 
    \vspace{-.1in}
     \caption{Learning curves for Experiment $\displaystyle 2$ with varying $pace(t) = \floor{kN}$ for \textit{DCL+}. The parameter $\displaystyle k$ needs to be finely tuned for improving the generalization of the network. A low $\displaystyle k$ value exposes only examples with less/no gradient noise to the network at every epoch whereas a high $\displaystyle k$ value exposes most of the dataset including examples with high gradient noise to the network. A moderate $\displaystyle k$ value shows examples with low/moderate gradient noise. Here, a moderate $\displaystyle k = 0.6$ generalizes the best.}
     \label{fig:exp2tunek}
     \vspace{-.1in}
     
 \end{figure}

In our experiments, we set $ ~\displaystyle pace(t) = \floor{kN} ~\forall t$, where $\displaystyle ~k \in [b/N,1] $ is a tunable hyperparameter. We use FCN-$10$ architecture to empirically validate our algorithms ($\displaystyle k = 0.9$) on a subset of the MNIST dataset with class labels $\displaystyle 0$ and $\displaystyle 1$ (Experiment 1). Since, this is a very easy task (as the \textit{vanilla} model training accuracy is as high as $\displaystyle \sim 99.9\%$), we compare the test loss values across training steps in Figure \ref{fig:exp1} to see the behaviour of DCL on an easy task. \textit{DCL+} shows the fastest convergence, although all the networks achieve the same test accuracy. \textit{DCL+} achieves \textit{vanilla}'s final (at training step $1000$) test loss score at training step $682$. 

In Experiment 2, we use FCN-$128$ to evaluate our DCL algorithms ($\displaystyle k = 0.6$) on a relatively difficult  Small Mammals dataset. Figure \ref{fig:exp2} shows that \textit{DCL+} achieves a faster and better convergence than \textit{vanilla} in Experiment 2. \textit{DCL+} achieves \textit{vanilla}'s convergence (at training step $4900$) test accuracy score at training step $1896$. Further experimental details are deferred to Suppl. \ref{app:networkdetails}.

Since, DCL is computationally expensive, we perform DCL experiments only on small datasets. Fine-tuning of $\displaystyle k$ is crucial for improving the generalization of \textit{DCL+} on the test set (see Figure \ref{fig:exp2tunek}). We fine-tune $\displaystyle k$ by trial-and-error over the training accuracy score. 
 
  \begin{figure*}[t]
 \centering
 \includegraphics[width=0.8\textwidth]{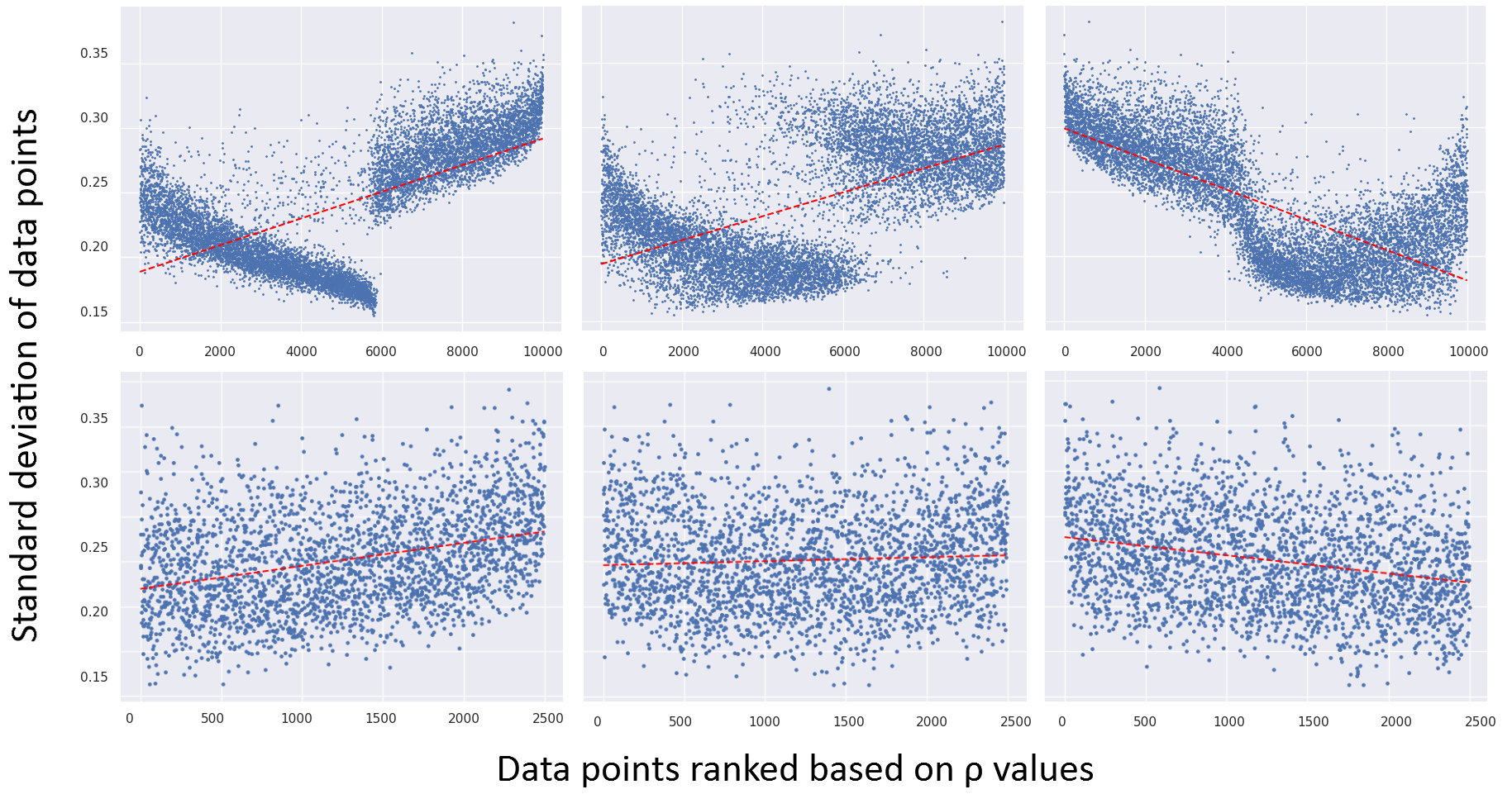}
 
     \caption{Relation of $\displaystyle \rho$ and $\displaystyle stddev$ values of examples over  training epochs 1 (left), 5 (middle), and 100 (right) for Experiments $\displaystyle 1$ (top row) and $\displaystyle 2$ (bottom row), respectively. Dotted red lines fit the scattered points.}
     \label{fig:stdrho}
     \vspace{-.1in}
     
 \end{figure*}
 
\section{Why is a curriculum useful?} \label{sec:whycl}
 
At an intuitive level, we can say that \textit{DCL+} converges faster than the \textit{vanilla} SGD as we greedily sample those examples whose gradient steps are the most aligned towards an approximate optimal weight vector. In previous CL works, mini-batches are generated by uniformly sampling examples from a partition of the dataset which is made by putting a threshold on the difficulty scores of the examples. Notice that our DCL algorithms generate mini-batches with a natural ordering at every epoch. We design \textit{DCL+} and \textit{DCL-} to investigate an important question: can CL benefit from having a set of mini-batches with a specific order or is it just the subset of data that is exposed to the learner that matters?  Figure \ref{fig:dcl} shows that the ordering of mini-batches matters while comparing \textit{DCL+} and \textit{DCL-}, which expose the same set of examples to the learner in any training epoch. Once the mini-batch sequence for an epoch is computed, \textit{DCL-} provides mini-batches to the learner in the decreasing order of gradient noise. This is the reason for \textit{DCL-} to have high discontinuities in the test loss curve after every epoch in Figure \ref{fig:exp1}. With our empirical results, we argue that the \textbf{ordering of mini-batches within an epoch does matter}.
 
\citet{cl} illustrates that removing examples that are misclassified by a Bayes classifier (\textit{noisy} examples) provides a good curriculum for training networks. SPL tries to remove examples that might be misclassified during a training step by avoiding examples with high loss. \textit{TL} avoids examples that are noisy to an approximate optimal hypotheses in the initial phases of training. \textit{DCL+} and \textit{DCL-} try to avoid examples with \textit{noisy gradients} that might slow down the convergence towards the desired optimal minima. \citet{cnet} empirically shows that avoiding examples with label noise improves the initial learning of CNNs. According to their work, adding examples with label noise to later phases of training serves as a regularizer and improves the generalization capability of CNNs. \textit{DCL+} uses its pace function to avoid highly noisy examples (in terms of gradients). In our DCL experiments, the parameter $\displaystyle k$ is chosen such that few moderately noisy examples (examples present in the last few mini-batches within an epoch) are included in training along with lesser noisy examples to improve the network's generalization. We also show the importance of tuning CL hyperparameters for achieving a better network generalization (see Figure \ref{fig:exp2tunek}). Hence, the parameter $\displaystyle k$ in \textit{DCL+} \textbf{serves as a regularizer and helps in improving the generalization of networks}.

\subsection{Analyzing \textit{stddev} with our DCL framework}
\label{analyzingstddevwithourdcl}

We use our DCL framework to understand why \textit{stddev} works as a scoring function. We try to analyze the relation between the standard deviation and $\displaystyle \rho_{t,i}$ values of examples over training epochs. Figure \ref{fig:stdrho} shows the plots of $stddev$ on the Y-axis against examples ranked based on their $\displaystyle \rho_{t,i}$ values (in ascending order) plotted on the  X-axis at various stages of training. It shows the dynamics of $\displaystyle \rho_{t,i}$ over initial, intermediate, and final stages of training. Correlation between $\rho_{t,i}$ and $stddev$ after the first epoch for Experiments 1 and 2 are $0.74$ and $0.36$, respectively. The corresponding p-values for testing non-correlation are $0$ and $3\times10^{-79}$, respectively. In the initial stage of training, examples with high $stddev$ tend to have high $\rho$ values. In the final stage of training, this trend changes to the exact opposite. This shows that $stddev$ can be useful in removing noisy gradients from the initial phases of training and hence help in defining a simple, good curriculum.

\section{Conclusion}

In this paper, we propose two novel CL algorithms that show improvements in network generalization over multiple image classification tasks with CNNs and FCNs. A fresh approach to define curricula for image classification tasks based on statistical measures is introduced, based on our observations from implicit curricula ordering. This technique makes it easy to score examples in an unsupervised manner without the aid of any teacher network. We thoroughly evaluate our CL algorithms and find it beneficial in noisy settings and improving network accuracy. We also propose a novel DCL algorithm for analyzing CL. We show that the ordering of mini-batches within training epochs and fine-tuning of CL hyperparameters are important to achieve good results with CL.  Further, we also use our DCL framework to support our CL algorithm that uses $stddev$ for scoring examples.

\bibliography{example_paper}
\bibliographystyle{icml2021}

\clearpage

\appendix

\section*{Supplementary Material}
\section{Additional empirical results} \label{app:aboutnorms}

 \begin{figure}[b!]
 \centering
    \includegraphics[width=0.35\textwidth]{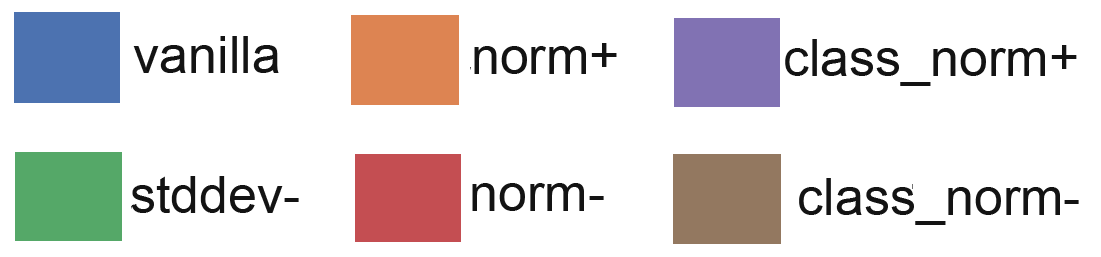}\label{fig:noisebarlegendsuppl}
     \subfigure[CIFAR-100]{\includegraphics[width=0.2\textwidth]{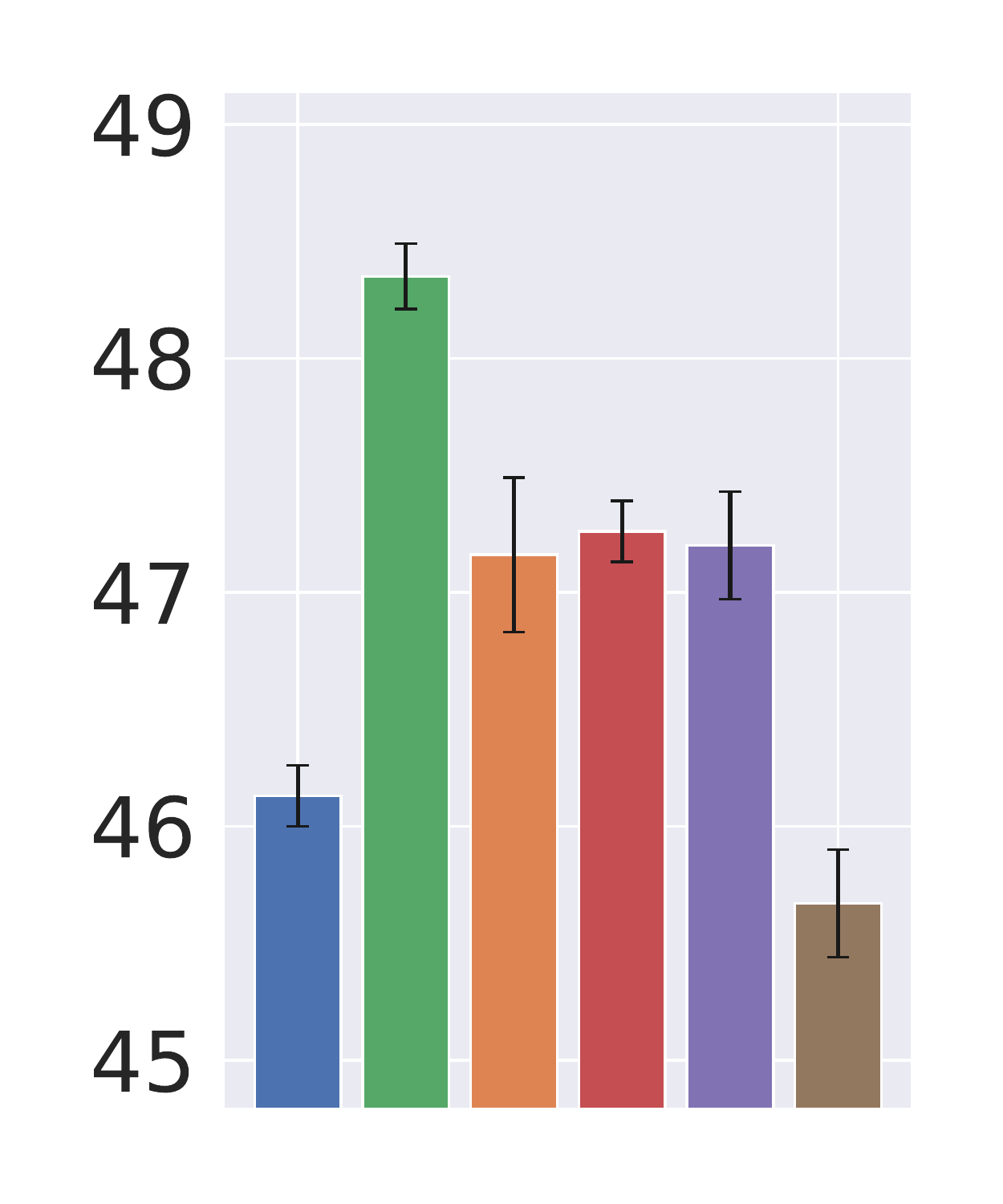}\label{fig:noisebarc100suppl}} 
     \subfigure[CIFAR-10]{\includegraphics[width=0.2\textwidth]{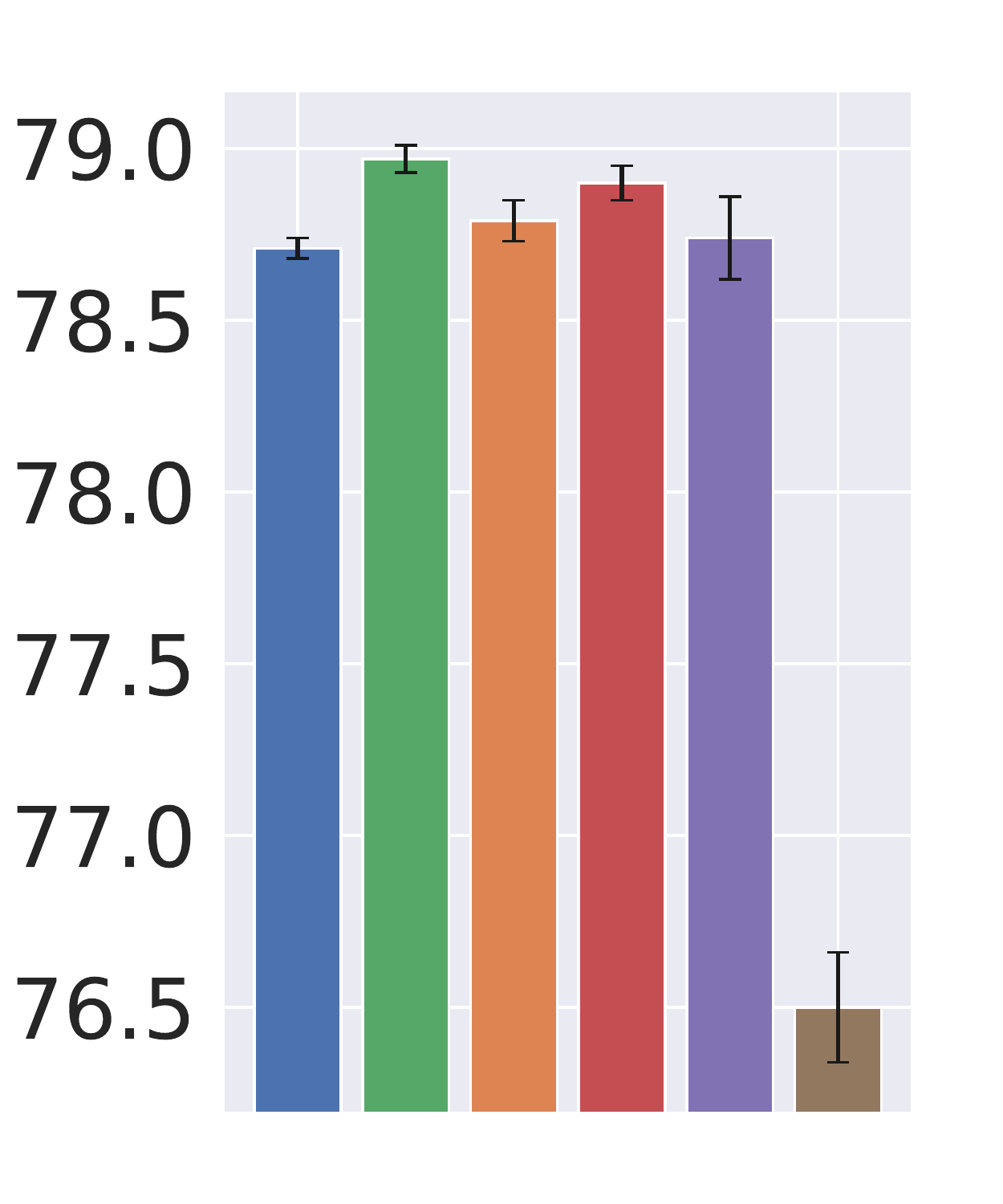}\label{fig:noisebarc10suppl}}
     \caption{Bars represent the final mean top-1 test accuracy (in $\displaystyle \%$) achieved by CNN-$8$. Error bars represent the STE after 25 independent trials.}
     \label{fig:normcompare}
     \vspace{-.1in}
     
 \end{figure}

In Section \ref{statisticalmeasuresfordefiningcurricula}, we study the performance of CL using $\displaystyle stddev$ and $\displaystyle entropy$ as scoring measures. Other important statistical measures are mode, median, and norm \citep{impstatmeasure}. A high $stddev$ for a real image could mean that the image is having a lot of edges and a wide range of colors. A low entropy could mean that an image is less noisy. Norm of an image could give information about its brightness. Intuitively, norm is not a good measure for scoring images as low norm valued images are really dark and high norm valued images are really bright. We experiment with different norm measures and find that they do not serve as a good CL scoring measure since they have lesser improvement with high variance over multiple trials when compared to \textit{stddev-} on the CIFAR datasets. We use two norm measures:

\begin{equation}
\begin{split}  \label{norms_score}
\displaystyle
norm(\vx) &= \|\vx\|_2 ~~~~~~~~\textrm{and} \\
class\_norm(\vx) &= \|\vx - \mu_\vx \|_2,
\end{split}
\end{equation}

where $\displaystyle \vx$ is an image in the dataset represented as a vector, and $\displaystyle \mu_\vx$ is the mean pixel value of all the images belonging to the class of $\displaystyle \vx$. In our experiments, all the orderings are performed based on the scoring function and the examples are then arranged to avoid class imbalance within a mini-batch. Let us denote the models that use the scoring functions $\displaystyle norm$ as \textit{norm+}, $\displaystyle -norm$ as \textit{norm-}, $\displaystyle class\_norm$ as \textit{class\_norm+}, and $\displaystyle -class\_norm$ as \textit{class\_norm-}.
 
Figure \ref{fig:normcompare} shows the results of our experiments on CIFAR-100 and CIFAR-10 datasets with CNN-$\displaystyle 8$ using $\displaystyle norm ~\textrm{and}~class\_norm$ scoring functions. We find that the improvements reported for \textit{norm-}, the best model among the models that use norm measures, have a lower improvement than \textit{stddev-}. Also, \textit{norm-} has a higher STE when compared to both \textit{vanilla} and \textit{stddev-}. Hence, based on our results, we suggest that $stddev$ is a more useful statistical measure  than norm measures for defining curricula for image classification tasks.

\section{Experimental Details} \label{app:expdetails}

\subsection{Network architectures} \label{app:networkdetails}

All FCNs (denoted as FCN-$m$) we use are $\displaystyle 2$-layered with a hidden layer consisting of $m$ neurons with ELU nonlinearities. Experiment 1 employs FCN-$\displaystyle 10$ while Experiment 2 employs FCN-$\displaystyle 128$ with no bias parameters. The outputs from the last layer is fed into a softmax layer. Cases 6--8 employ FCN-$\displaystyle 512$ with bias parameters. The batch-size for Experiments 1--2 and Cases 1--9 are 50 and 100, respectively. We use one NVIDIA Quadro RTX 5000 GPU for our experiments. Average runtimes of our experiments vary from 1 hour to 3 days.

For Cases 1--5 and 9, we use the CNN-$8$ architecture that is used in \citet{powerofcl}. The codes are available in their GitHub repository. CNN-$8$ contains $8$ convolution layers with $32, 32, 64, 64, 128, 128, 256,$ and $256$ filters, respectively, and ELU nonlinearities. Except for the last two convolution layers with filter size $\displaystyle 2\times2$, all other layers have a filter size of $\displaystyle 3\times3$. Batch normalization is performed after every convolution layer. $\displaystyle 2\times2$ max-pooling and $\displaystyle 0.25$ dropout layers are present after every two convolution layers. The output from the CNN is flattened and fed into a fully-connected layer with $\displaystyle 512$ neurons followed by a $\displaystyle 0.5$ dropout layer. A softmax layer follows the fully-connected output layer that has a number of neurons same as the number of classes in the dataset. The batch-size is $\displaystyle 100$. All the CNNs and FCNs are trained using SGD with cross-entropy loss. SGD uses an exponential step-decay learning rate scheduler. Our codes will be published on acceptance.

\subsection{Hyperparameter tuning} \label{app:tuning}

For fair comparison of network generalization, the hyperparameters should be finely tuned as mentioned in \citet{powerofcl}. We exploit hyperparameter grid-search to tune the hyperparameters of the models in our experiments. For \textit{vanilla} models, grid-search is easier since they do not have a pace function. For CL models, we follow a coarse two-step tuning process as they have a lot of hyperparameters. First we tune the optimizer hyperparameters fixing other CL hyperparameters. Then we fix the obtained optimizer parameters and tune the CL hyperparameters.

The gird-search parameter ranges are as follows. Case 1: a) initial learning rate $0.01-0.1$ b) learning rate exponential decay factor $1.1-2$ c) learning rate decay step $200-800$ d) $step\_length$ $20-400$ e) $inc$ $1.1-3$ f) $starting\_fraction$ $0.04-0.15$. Cases 2--3, 9: a) initial learning rate $0.05-0.2$ b) learning rate exponential decay factor $1.1-2$ c) learning rate decay step $200-800$ d) $step\_length$ $100-2000$ e) $inc$ $1.1-3$ f) $starting\_fraction$ $0.04-0.15$. Case 4--5, 9: a) initial learning rate $0.005-0.5$ b) learning rate exponential decay factor $2-10$ c) learning rate decay step $100-12000$ d) $step\_length$ $10-100$ e) $inc$ $1.1-2$ f) $starting\_fraction$ $0.04-0.15$. Cases 6--8: a) initial learning rate $0.001-0.01$ b) learning rate exponential decay factor $1.1-2$ c) learning rate decay step $200-800$ d) $step\_length$ $20-100$ e) $inc$ $1.1-2$ f) $starting\_fraction$ $0.04-0.15$. The experiments are tuned to perform better on the training data.

\subsection{Dataset details} \label{app:dataset}

We use CIFAR-100, CIFAR-10, ImageNet Cats, Small Mammals, MNIST, and Fashion-MNIST datasets. CIFAR-100 and CIFAR-10 contain $\displaystyle 50,000$ training and $\displaystyle 10,000$ test images of shape $\displaystyle 32\times32\times3$ belonging to $\displaystyle 100$ and $\displaystyle 10$ classes, respectively. Small Mammals is a super-class of CIFAR-100 containing 5 classes -- ``Hamster'', ``Mouse'', ``Rabbit'', ``Shrew'', and ``Squirrel''. It has $\displaystyle 500$ training images per class and $\displaystyle 100$ test images per class. MNIST and Fashion-MNIST contain $\displaystyle 60,000$ training and $\displaystyle 10,000$ test gray-scale images of shape $\displaystyle 28\times28$ belonging to $\displaystyle 10$ different classes. ImageNet Cats is a subset of the ImageNet dataset ILSVRC 2012. It has $7$ classes with each class containing $1300$ training images and $50$ test images. The labels in the subset are ``Tiger cat'', ``Lesser panda, Red panda, Panda, Bear cat, Cat bear, Ailurus fulgens'', ``Egyptian cat'', ``Persian cat'', ``Tabby, Tabby cat'', ``Siamese cat, Siamese'', ``Madagascar cat'', and  ``Ring-tailed lemur, Lemur catta''. The images in the dataset are reshaped to $56\times56\times3$. All the datasets are preprocessed before training to have a zero mean and unit standard deviation.

\end{document}